\documentclass[11pt,a4paper]{article}
\usepackage[hyperref]{emnlp2020}
\usepackage{times}
\usepackage{latexsym}

\usepackage{microtype}

\aclfinalcopy 
\def\aclpaperid{1219} 

\usepackage{graphicx}
\usepackage{multirow}
\usepackage{array}
\usepackage{booktabs}
\usepackage{subcaption}

\title{Incremental Processing in the Age of Non-Incremental Encoders: An Empirical Assessment of Bidirectional Models for Incremental NLU}

\author{Brielen Madureira \hspace{1cm} David Schlangen \\
	Computational Linguistics \\
	Department of Linguistics \\
	University of Potsdam, Germany \\
	\texttt{\{madureiralasota, david.schlangen\}@uni-potsdam.de} \\
}

\date{}

\begin{document}
\maketitle
\begin{abstract}
  While humans process language incrementally, the best language encoders currently used in NLP do not. Both bidirectional LSTMs and Transformers assume that the sequence that is to be encoded is available in full, to be processed either forwards and backwards (BiLSTMs) or as a whole (Transformers). We investigate how they behave under incremental interfaces, when partial output must be provided based on partial input seen up to a certain time step, which may happen in interactive systems. We test five models on various NLU datasets and compare their performance using three incremental evaluation metrics. The results support the possibility of using bidirectional encoders in incremental mode while retaining most of their non-incremental quality. The ``omni-directional'' BERT model, which achieves better non-incremental performance, is impacted more by the incremental access.
  This can be alleviated by adapting the training regime (truncated training), or the testing procedure, by delaying the output until some right context is available or by incorporating hypothetical right contexts generated by a language model like GPT-2. 
\end{abstract}

\section{Introduction}
\label{intro}
In ``The Story of Your Life'', a science fiction short story by Ted~\citet{chiangstory}, 
Earth is visited by alien creatures whose writing system does not unfold in time but rather presents full thoughts instantaneously. In our world, however, language \emph{does} unfold over time, both in speaking and in writing. There is ample evidence~\citep[\textit{inter alia}]{marslenwilson:1975,tanenhaus2008language} that it is also \emph{processed} over time by humans, in an incremental fashion where the interpretation of a full utterance is continuously built up while the utterance is being perceived.

In Computational Linguistics and Natural Language Processing, this property is typically abstracted away by assuming that the unit to be processed (\textit{e.g.}, a sentence) is available as a whole.\footnote{%
	An exception is the field of research on interactive systems, where it has been shown that incremental processing can lead to preferable timing behavior~\citep{Aistetal:incrunder,skantze2009incremental} and work on incremental processing is ongoing~\citep[\textit{inter alia}]{lectrack2015,trinh2018multi,coman2019}.
} The return and subsequent mainstreaming of Recurrent Neural Networks (RNNs), originally introduced by~\citet{elman1990finding} and repopularized \emph{i.a.}\ by~\citet{mikolov2010recurrent}, may have made it seem that time had found a place as a first-class citizen in NLP. However, it was quickly discovered that certain technical issues of this type of model could be overcome, for example in the application of machine translation, by encoding input sequences in reverse temporal order~\citep{sutskever2014sequence}. 

This turns out to be a special case of the more general strategy of bidirectional processing, proposed earlier in the form of BiRNNs~\citep{schuster1997bidirectional,baldi1999exploiting} and~BiLSTMs~\citep{hochreiter1997long}, which combine a forward and a backward pass over a sequence. More recently, Transformers \citep{vaswani2017attention} also function with representations that inherently have no notion of linear order. Atemporal processing has thus become the standard again.

In this paper, we explore whether we can adapt such bidirectional models to work in incremental processing mode and what the performance cost is of doing so. We first go back and reproduce the work of~\citet{huang2015bidirectional}, who compare the performance of LSTMs and BiLSTMs in sequence tagging, extending it with a BERT-based encoder and with a collection of different datasets for tagging and classification tasks. Then we address the following questions:
\vspace{0.3cm}

\begin{figure*}[h]
	\center
	\includegraphics[trim={0.5cm 19cm 16cm 0.5cm},clip, width=11cm]{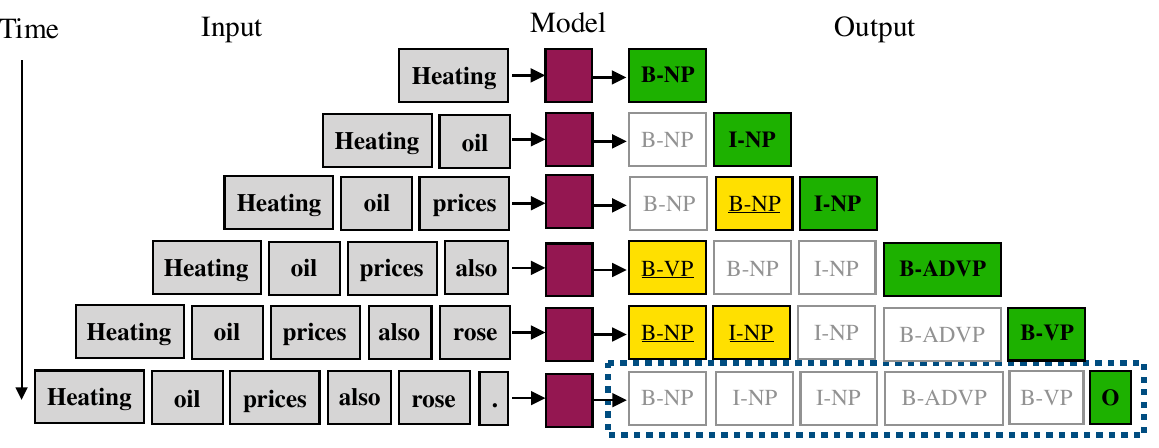}
	\caption{Incremental interface on a bidirectional tagging model (here for chunking). Each line represents the input and output at a time step. Necessary additions are green/bold, substitutions are yellow/underlined, and the dashed frame shows the output of the final time step, which is the same as the non-incremental model's.}
	\label{fig:incremental}
\end{figure*}

\textbf{Q1}. \textbf{If we employ inherently non-incremental models in an incremental system, do we get functional representations that are adequate to build correct and stable output along the way?} We examine how bidirectional encoders behave under an incremental interface, revisiting the approach proposed by~\citet{beuck2011decision} for POS taggers. After standard training, we modify the testing procedure by allowing the system to see only successively extended prefixes of the input available so far with which they must produce successively extended prefixes of the output, as shown in Figure~\ref{fig:incremental}. The evaluation metrics are described in Section~\ref{evalmetrics}, and the discussion is anchored on the concepts of timeliness, monotonicity, and decisiveness and their trade-off with respect to the non-incremental quality~\citep{beuck2011decision,kohn2018incremental}. We show that it is possible to use them as components of an incremental system (\textit{e.g.} for NLU) with some trade-offs.

\textbf{Q2}. \textbf{How can we adapt the training regime or the real-time procedure to mitigate the negative effect that the non-availability of right context (i.e., future parts of the signal) has on non-incremental models?} To tackle this question, we implement three strategies that help improve the models' incremental quality: \textit{truncated training}, \textit{delayed output} and \textit{prophecies} (see Section~\ref{framework}).

Our results are relevant for incremental Natural Language Understanding, needed for the design of dialogue systems and more generally interactive systems, \textit{e.g.} those following the incremental processing model proposed by~\citet{schlangen2011general}. These systems rely on the availability of partial results, on which fast decisions can be based. Similarly, simultaneous translation is an area where decisions need to be based on partial input with incomplete syntactical and semantic information.

\section{Related Work}
\label{litreview}

\subsection{Bidirectionality}

Language is one of the cognitive abilities that have a temporal nature. The inaugural adoption of~RNNs~\citep{elman1990finding} in NLP showed a pursuit to provide connectionist models with a dynamic memory in order to incorporate time implicitly, not as a dimension but through its effects on processing. Since then, the field has witnessed the emergence of a miscellany of neural architectures that take the temporal structure of language into account. In particular,~LSTMs ~\citep{hochreiter1997long} have been vastly used for sequence-to-sequence or sequence classification tasks, which are ubiquitous in~NLP.

Bidirectional~LSTMs~\citep{schuster1997bidirectional,baldi1999exploiting} are an extension to~LSTMs that exploit bidirectionality and whose basic processing units are full sentences. They achieved remarkable results in many tasks, \textit{e.g.} part-of-speech tagging~\citep{ling2015finding,plank2016multilingual}, chunking~\citep{zhai2017neural}, named entity recognition~\citep{chiu2016named}, semantic role labeling~\citep{he2017deep}, slot filling and intent detection~\citep{haihong2019novel} and opinion mining~\citep{irsoy2014opinion}. Subsequent works have confirmed that bidirectionality can afford an increase in performance~\citep{graves2005framewise,huang2015bidirectional,zhai2017neural}.

More recently,~\citet{vaswani2017attention} has consolidated the application of attention mechanisms on NLP tasks with Transformers, which are not constrained by only two directions, as~BiLSTMs. Instead, complete sentences are accessed at once. 
The need for NLP neural networks to be grounded on robust language models and reliable word representations has become clear. The full right and left context of words started to play a major role as in~\citet{peters2018elmo}, which resorts to bidirectionality to train a language model. In a combination of bidirectional word representations with the Transformer architecture, we observe the establishment of~BERT~\citep{devlin2019bert} as a current state-of-the-art model, on top of which an output layer can be added to solve classification and tagging tasks.

\subsection{Incremental processing}

The motivation to build incremental processors, as defined by ~\citet{kempen1982incremental} and~\citet{levelt1989}, is twofold: they are more cognitively plausible and, from the viewpoint of engineering, real-time applications such as parsing~\citep{nivre-2004-incrementality},~SRL~\citep{konstas2014incremental},~NLU~\citep{peldszus2012joint}, dialog state tracking~\citep{trinh2018multi},~NLG and speech synthesis~\citep{buschmeier-etal-2012-combining} and~ASR~\citep{selfridge-etal-2011-stability} require that the input be continually evaluated based on incoming prefixes while the output is being produced and updated. 

Another advantage is a better use of computational resources, as a module does not have to wait for the completion of another one to start processing~\citep{skantze2009incremental}. In robots, linguistic processing must also be intertwined with its perceptions and actions, happening simultaneously~\citep{brick2007incremental}. 

Research on processing and generating language incrementally has been done long before the current wave of neural network models, using several different methods. For example, in ASR, a common strategy has been to process the input incrementally to produce some initial output, which was then re-scored or re-processed with a more complex model~\citep{vergyri2003prosodic,hwang2009building}. While the recent accomplishments of neural encoders are cherished, bidirectional encoders drift apart from a desirable temporal incremental approach because they are trained to learn from complete sequences.

There is some cognitive resemblance underlying~RNNs in the sense that they can process sequences word-by-word and build intermediary representations at every time step. This feature provides a legitimate way to employ them in incremental systems.~\citet{trinh2018multi} and~\citet{lectrack2015} explore this, for instance, using the~LSTM's representations to predict dialogue states after each word. Recent works on simultaneous translation also use RNNs as incremental decoders~\citep{dalvi-etal-2018-incremental}.

Some works arouse interest in the incremental abilities of RNNs.~\citet{hupkes2018} use a diagnostic classifier to analyze the representations that are incrementally built by sequence-to-sequence models in disfluency detection and conclude that the semantic information is only kept encoded for a few steps after it appears in the dialogue, being soon forgotten afterwards.~\citet{ulmer2019} propose three metrics to assess the incremental encoding abilities of~LSTMs and compare it with the addition of attention mechanisms.

According to~\citet{beuck2011decision} and~\citet{schlangen2011general}, incrementality is not a binary feature. Besides using inherently incremental algorithms, it is also possible to provide incremental \emph{interfaces} to non-incremental algorithms. Such interfaces simply feed ever-increasing prefixes to what remains a non-incremental algorithm, providing some ``housekeeping'' to manage the potentially non-monotonic results.

To alleviate the effect of the partiality of the input, we test the use of anticipated continuations, inspired by the mechanism of predictive processing discussed in cognitive science~\citep{christiansen2016} and the idea of interactive utterance completion introduced by~\citet{devault2011incremental}. Related strategies to predict upcoming content and to wait for more right context are also applied in recent work on simultaneous translation~\citep{grissom-rl-simul-mt,oda-etal-2015-syntax,ma-etal-2019-stacl}. The use of truncated inputs during training, discussed below, aims at making intermediate structures available during learning, an issue discussed in~\citet{kohn2018incremental}. This is a variation of chunked training used in~\citet{dalvi-etal-2018-incremental}.

\section{Evaluation of incremental processors}
\label{evalmetrics}

The hierarchical nature of language makes it likely that incremental processing leads to non-monotonic output due to re-analysis, as in the well-known ``garden path'' sentences.
Incremental systems may edit the output by adding, revoking, and substituting its parts~\citep{baumann2011evaluation}. We expect an incremental system to produce accurate output as soon as possible~\citep{trinh2018multi}, with a minimum amount of revocations and substitutions, ideally only having correct additions, to avoid jittering that may be detrimental to subsequent processors working on partial outputs. 

To assess the incremental behavior of sequence tagging and classification models, we use the evaluation metrics for incremental processors established by~\citet{schlangen2009incremental} and~\citet{baumann2011evaluation}. The latter defines three diachronic metrics: \textit{edit overhead} (EO $\in [0,1]$), the proportion of unnecessary edits (the closer to 0, the fewer edits were made); \textit{correction time} (CT $\in [0,1]$), the fraction of the utterance seen before the system commits on a final decision for a piece of the output (the closer to 0, the sooner final decisions were made); and \textit{relative correctness} (RC $\in [0,1]$), the proportion of outputs that are correct with respect to the non-incremental output (being close to 1 means the system outputs were most of the time correct prefixes of the non-incremental output). 

The sequence tagging tasks we evaluate are massively incremental~\citep{hildebrandt1999inkrementelle}, meaning that a new label is always added to the output after a new word is processed. The models can also substitute any previous labels in the output sequence in the light of new input. Sequence classifiers must add one label (the sequence's class) after seeing the first word and can only substitute that single label after each new word. In both cases, additions are obligatory and substitutions should ideally be kept as low as possible, but there can be no revocations. Moreover, our data is sequential, discrete, and order-preserving \cite{kohn2018incremental}. 

Given a sequence of length $n$, the number of necessary edits is always the number of tokens in the sequence (all additions) for sequence taggers and we set it to 1 for sequence classifiers. All other edits (substitutions) count as unnecessary and their number is bounded by $\sum_{i=1}^{n-1}i$ 
for tagging, and by $(n-1)$, for classification. 

\begin{figure}
	\centering
	\includegraphics[trim={0cm 18cm 21cm 0cm},clip, width=\linewidth]{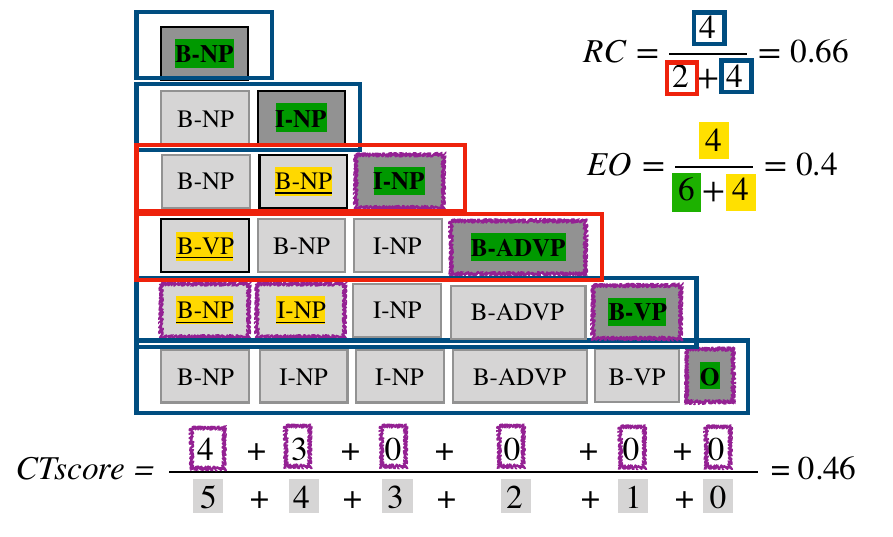}
	\caption{How we estimate the evaluation metrics for the complete sequence of outputs from Figure~\ref{fig:incremental}.}
	\label{fig:metrics}
\end{figure}

We need to slightly adapt the CT measure for sequences. It is originally defined as FD-F0, the time step of a final decision minus the time step when the output first appeared. F0 is fixed for every word in a sequence (the systems always output a new label corresponding to each new word it sees), but each label will have a different FD. In order not to penalize initial labels, which have more opportunities of being substituted than final ones, we instead sum the FD of each token and divide by the sum of the number of times each one could be modified, to get a score for the sequence as a whole. Let the sequence length be $n$, then here CTscore $=(\sum_{i=1}^{n} F\!D_i) / (\sum_{i=1}^{n} n-i)$. We define it to be 0 for sequences of one token.  Again, 0 means every label is immediately committed, 1 means all final decisions are delayed until the last time step. Figure \ref{fig:metrics} presents a concrete example of how to estimate the metrics. 

Based on the trade-off between responsiveness and output quality~\citep{skantze2009incremental}, we also estimate whether there is any improvement in the quality of the outputs if the encoder waits for some right context to appear before committing on output previously generated. For that, we use delayed EO and delayed RC (also named \emph{discounted} in~\citealp{baumann2011evaluation}), which allows one or two words of the right context to be observed before outputting previous labels, named EO/RC$\Delta1$ and EO/RC$\Delta2$, respectively. 

In order to concentrate on the incremental quality despite the eventual non-incremental deficiencies, we follow the approach by~\citet{baumann2011evaluation} and evaluate intermediate outputs in comparison to the processor's final output, which may differ from the gold output but is the same as the non-incremental output. The general non-incremental correctness should be guaranteed by having high accuracy or F1 score in the non-incremental performance. 

\section{Models}
\label{framework}

\begin{figure}[h]
	\centering
	\includegraphics[trim={0cm 16cm 23cm 2.3cm},clip, width=\linewidth]{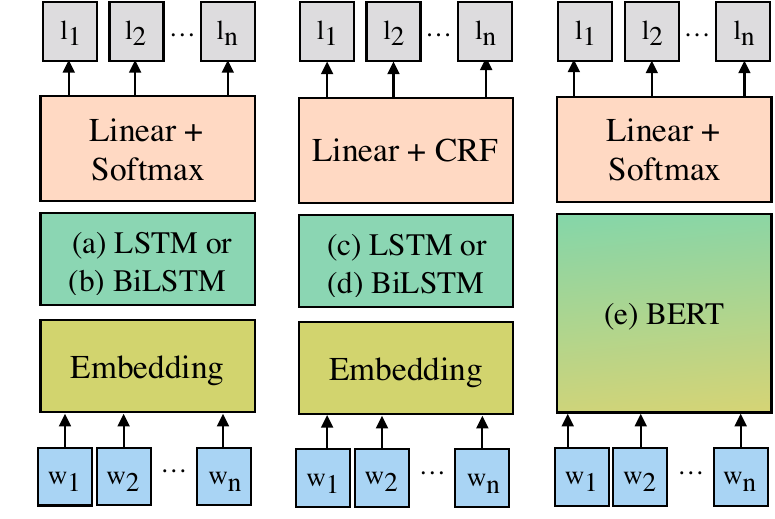}
	\caption{Models for sequence tagging, w=word and l=label. (a) is the only inherently incremental. (a), (b) and (e) can also be used for sequence classification if we consider only their final representation.}
	\label{fig:models}
\end{figure}

We test the behavior of five neural networks, illustrated in Figure~\ref{fig:models}, under an incremental processing interface operating on word level and having full sentences as processing units: a) a vanilla~LSTM; b) a vanilla~BiLSTM; c) an~LSTM with a~CRF (Conditional Random Field) layer; d) a~BiLSTM with a~CRF layer; and e)~BERT. The vanilla~LSTM is the only model that works solely in temporal direction.

We choose to use the basic forms of each model to isolate the effect of bidirectionality. They perform well enough on the tasks to enable a realistic evaluation (see Table~\ref{table:performance}). Note that state-of-the-art results are typically achieved by combining them with more sophisticated mechanisms.

We use the models for both sequence tagging and classification. They use the representation at each time step to predict a corresponding label for sentence tagging, whereas for sequence classification they use the representation of the last time step~(LSTM) or a combination of the last forward and backward representations~(BiLSTM) or, in case of BERT, the representation at the~CLS (initial) token, as suggested in~\citet{devlin2019bert}. The two models with~CRF cannot be used for classification, as there are no transition probabilities to estimate.

Sequence tagging implies a one-to-one mapping from words to labels, so that for every new word the system receives, it outputs a sequence with one extra label. In sequence classification, we map every input to a single label. In that case, the~LSTM can also edit the output since it can change the chosen label as it processes more information. Because the datasets we use are tokenized and each token has a corresponding label, we follow the instructions given by~\citet{devlin2019bert} for dealing with~BERT's subtokenization: the scores of the first subtoken are used to predict its label, and further subtoken scores are ignored.

\begin{figure}
	\center
	\vspace{0.8cm}
	\includegraphics[trim={0 20cm 23cm 0.5cm},clip, width=\linewidth]{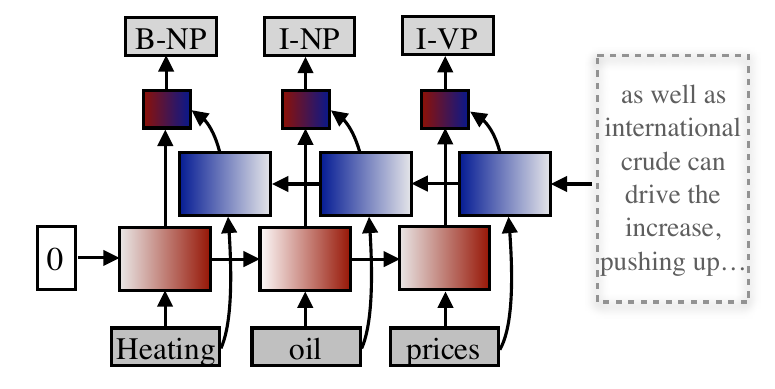}
	\caption{Incremental interface of a non-incremental bidirectional model, showing the input and output at time step 3. The context vector fed into the backward~LSTM can be zero or initialized with a hypothetical right context generated by a language model.}
	\label{fig:inc-interface}
\end{figure}

Except for the~LSTM on sequence tagging, all models' outputs are non-monotonic, i.e., they may reassign labels from previous words. The concept of timeliness is trivial here because we know exactly that the label for the \textit{t}-th word will appear for the first time at the \textit{t}-th version of the output, for all \textit{t}. Even so, we can delay the output to allow some lookahead. In terms of decisiveness, all models commit to a single output at every time step. 
Figure~\ref{fig:inc-interface} shows an example of the computation graph.~BiLSTMs can recompute only the backward pass, while~BERT needs a complete recomputation. 

\subsection{Strategies}

We check the effect of three strategies: \textit{truncated training}, \textit{delayed output} and \textit{prophecies}. In the first case, we modify the training regime by stripping off the endings of each sentence in the training set. 
We randomly sample a maximum length $l \leq n$, where $n$ is the original sentence length, and cut the subsequent words and labels. We expect this to encourage the model to know how to deal with truncated sequences that it will have to process during testing. 

\begin{figure}[h]
	\center
	\includegraphics[trim={2.4cm 20cm 22.5cm 0.3cm},clip, width=0.9\linewidth]{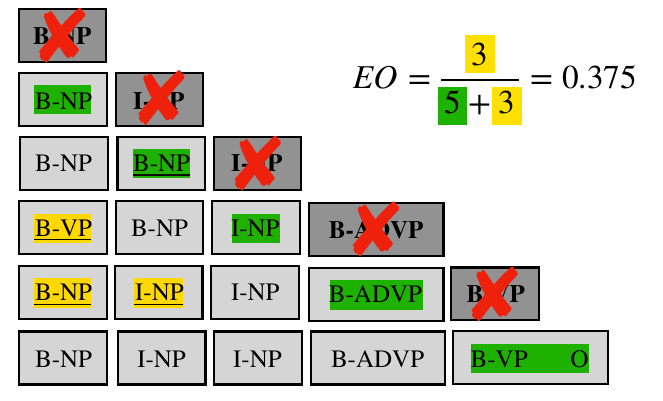}
	\caption{Example of the calculation of Edit Overhead with $\Delta1$ delay for the example in Figure \ref{fig:metrics}. The first choice for each label happens once the subsequent word has been observed, except for the last token in the sentence.}
	\label{fig:lookahead}
\end{figure}

The second strategy involves allowing some upcoming words to be observed before outputting a label corresponding to previous words. This is a case of lookahead described in ~\citet{baumann2011evaluation}, where the processor is allowed to wait for some right context before making a first decision with respect to previous time steps. We experiment with right contexts of one or two words, $\Delta1$ and $\Delta2$, respectively. $\Delta1$ means the model outputs the first label for word $t$ once it consumes word $t+1$. Analogously, $\Delta2$ means the model can observe words $t+1$ and $t+2$ before outputting the first label for word $t$. Figure \ref{fig:lookahead} illustrates how to calculate $EO$ with $\Delta1$ delay for the same example as in Figure \ref{fig:metrics}.

\begin{figure}[h]
	\center
	\includegraphics[trim={0cm 16.4cm 19cm 0.3cm},clip, width=\linewidth]{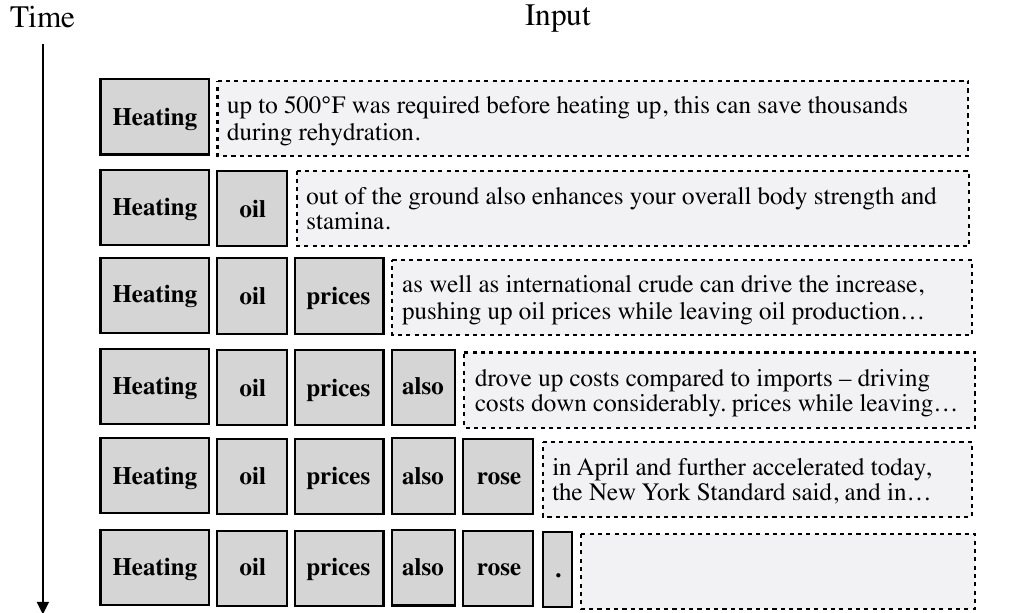}
	\caption{Input throughout time steps using hypothetical right contexts generated by~GPT-2, providing a full sequence for the backward direction.}
	\label{fig:incremental-prophecies}
\end{figure}

In the third strategy, we first feed each prefix as left context in the~GPT-2 language model and let it generate a continuation up to the end of a sentence to create a hypothetical full context that meets the needs of the non-incremental nature of the models (see Figure \ref{fig:incremental-prophecies} for an example). Not surprisingly, the mean BLEU scores of the prophecies with respect to the real continuation of the sentences are less than 0.004 for all datasets.\footnote{ Fine-tuning GPT-2 did not improve BLEU and caused marginal difference in the evaluation metrics. We thus present the results using the pre-trained model only, and leave more exploration of fine-tuning for future work.}

\section{Experiments}
\label{experiments}

\begin{table*}[!h]
	\small
	\centering
	
	\begin{tabular}{r c c c c c c} 
		\toprule 
		& & \multicolumn{5}{c}{\bf Model } \\ 
		\cmidrule{3-7} 
		\bf Task & \bf Metric & \bf LSTM & \bf LSTM+CRF & \bf BiLSTM & \bf BiLSTM+CRF & \bf BERT \\ 
		\cmidrule{1-7} 
		Chunk & & 86.93 (84.23) & 90.40 (88.13) & 90.22 (88.07) & 91.24 (89.44) & 96.32 (96.11)\\ 
		Named Entity Recognition & &  70.78 (67.64) & 86.30 (83.98) & 88.79 (84.72) & 89.29 (87.50) & 93.52 (92.54)\\ 
		Semantic Role Labeling & F1 Score & 52.27 (49.63) & 70.83 (68.71) & 77.39 (73.34) & 84.28 (80.88) & 89.01 (87.23)\\ 
		Slot Filling (ATIS) & (\%) & 93.82 (90.78) & 95.36 (92.09) & 94.84 (91.41) & 95.26 (92.63) & 95.57 (93.88)\\ 
		Slot Filling (SNIPS) & & 82.20 (78.09) & 89.63 (85.28) & 90.44 (85.41) & 92.32 (87.82) & 95.46 (92.93)\\ 
		\cmidrule{2-7} 
		Intent (ATIS) &  & 96.86 (93.06) & -  & 95.74 (93.62) & -  & 97.31 (95.86) \\ 
		Intent (SNIPS) & &96.86 (97.43) & -  & 97.43 (97.43) & -  & 97.57 (97.71) \\ 
		Part-of-Speech Tagging & Accuracy & 94.98 (94.32)  & 96.02 (95.56)  & 96.44 (96.23)  & 96.64 (96.35)  & 97.87 (97.65) \\ 
		Positive/Negative & (\%) & 82.17 (72.83) & -  & 83.33 (75.67) & -  & 93.83 (92.50) \\ 
		Pros/Cons & & 94.51 (93.85) & -  & 94.40 (93.65) & -  & 95.74 (95.17) \\ 
		\bottomrule 
	\end{tabular} 
	
	\caption{Non-incremental performance of all models on test sets (truncated training in parentheses). The results are not necessarily state-of-the-art because we use basic forms of each model in order to isolate the effect of bidirectionality and have comparable results among different tasks.}
	\label{table:performance}
\end{table*} 

\subsection{Data}

We examine the incremental evaluation metrics on ten datasets in English, six for sequence tagging: chunking~\citep{conllchunk}, slot filling~\citep[ATIS and SNIPS, respectively]{hemphill1990atis,coucke2018snips}, named entity recognition, part-of-speech tagging and semantic role labeling ~\citep{ontonotes}; and four for sentence classification: intent~\citep[ATIS and SNIPS, respectively]{hemphill1990atis,coucke2018snips} and sentiment \citep[positive/negative and pros/cons, respectively]{kotzias2015posneg,ganapathibhotla2008proscons}. 

Chunking,~NER,~SRL, and slot filling use the~BIO labeling scheme and are evaluated using the F1 score adapted for sequence evaluation, whereas the performance on~POS tagging and classification tasks is measured by accuracy.  

The models map from raw words to labels without using any intermediate annotated layer, even though they are available in some datasets. The only exception is the SRL task, for which we concatenate predicate embeddings to word embeddings following the procedure described in~\citet{he2017deep}, because a sequence can have as many label sequences as its number of predicates. 

\subsection{Implementation}

During training, we minimize cross entropy using the Adam method for optimization~\citep{kingma2014adam}.  We perform hyperparameter search for the~LSTM model using Comet's Bayes search algorithm,\footnote{\url{http://www.comet.ml}} to maximize the task's performance measure on the validation set and use its best hyperparameters for all other models, except ~BERT, for which we use HuggingFace's pre-trained bert-base-cased model.

We use GloVe embeddings~\citep{pennington2014glove} to initialize word embeddings for all models except~BERT, which uses its own embedding mechanism. Random embeddings are used for out-of-GloVe words. We randomly replace tokens by a general $<$unk$>$ token with probability 0.02 and use this token for all unknown words in the validation and test sets~\citep{lectrack2015}.

No parameters are kept frozen during training. Overfitting is avoided with early stopping and dropout. Our implementation uses PyTorch v.1.3.1, and prophecies are generated with~HuggingFace's port of the GPT-2 language model. The evaluation of incrementality metrics is done on the test sets. \footnote{The code is available at \url{https://github.com/briemadu/inc-bidirectional}. For more details on implementation and data for reproducibility, see Appendix.}

\section{Results}
\label{results}

The results in Table~\ref{table:performance} (above) support the observation that, in general,  bidirectional models do have a better non-incremental performance than~LSTMs (except for~IntentATIS and~ProsCons) and that there is an overall considerable improvement in the use of~BERT model for all tasks. Truncated training reduces overall performance but even so~BERT with truncated training outperforms all models, even with usual training, in most tasks (except for slot filling and~IntentATIS). 

\begin{figure}[h!]
	\begin{subfigure}[t]{\linewidth}
		\centering
		\includegraphics[trim={0cm 9.8cm 0cm 0cm}, clip, height=0.5cm]{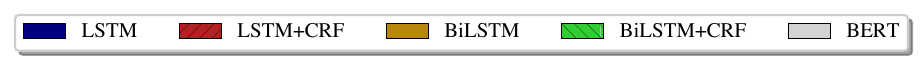} 
	\end{subfigure}
	
	\centering
	\begin{subfigure}[t]{\linewidth}
		\centering
		\includegraphics[trim={0.5cm 0.3cm 0.5cm 1cm}, clip, height=5cm]{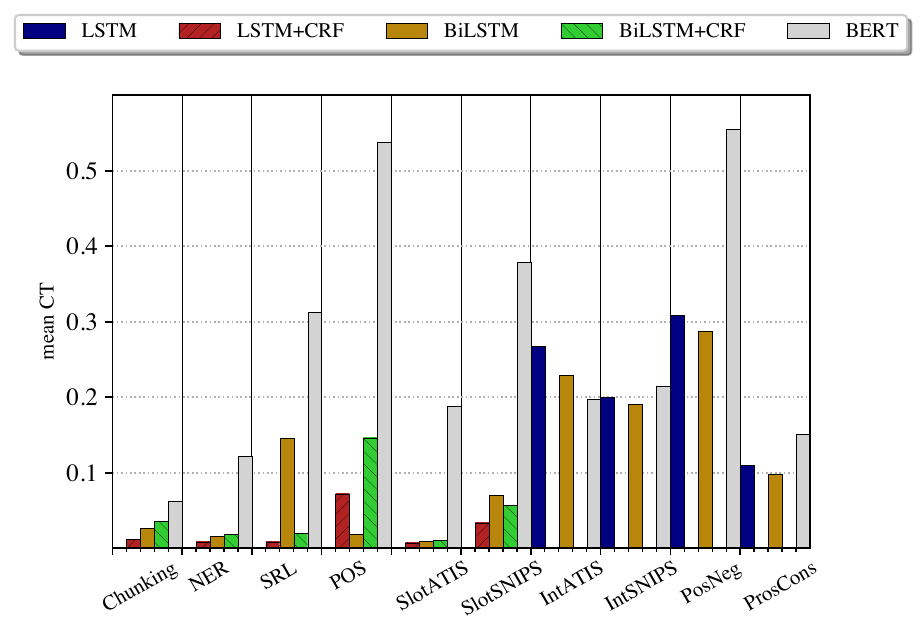} 
		\caption{Mean Correction Time Score}
		\label{fig:rose-ct}
	\end{subfigure}
	
	\centering
	\begin{subfigure}[t]{\linewidth}
		\centering
		\includegraphics[trim={0.5cm 0.3cm 0.5cm 1cm}, clip, height=5cm]{figures/columns_EO.pdf} 
		\caption{Mean Edit Overhead}
		\label{fig:rose-eo}
	\end{subfigure}
	
	\begin{subfigure}[t]{\linewidth}
		\centering
		\includegraphics[trim={0.5cm 0.3cm 0.5cm 1cm}, clip, height=5cm]{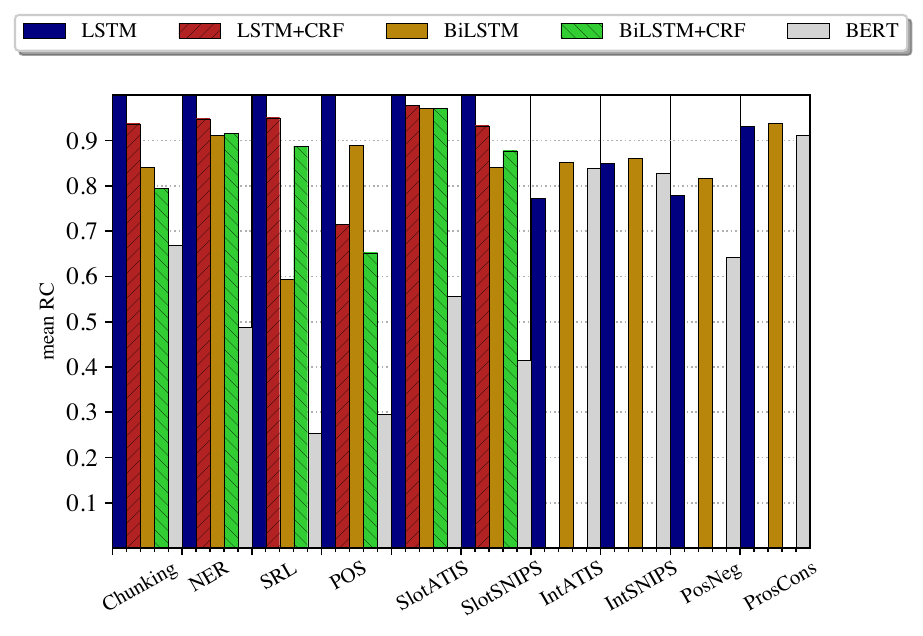} 
		\caption{Mean Relative-Correctness}
		\label{fig:rose-rc}
	\end{subfigure}
	
	\caption{Comparison of evaluation metrics for all models and tasks. The incremental behavior is more stable for sequence tagging than for sequence classification.~BERT takes longer to reach final decisions, and its outputs are edited more often than other models, especially in sequence tagging tasks.}
	\label{fig:roses}
	
\end{figure}

Figure~\ref{fig:roses} presents an overview of the incremental evaluation metrics for all models and tasks. Sequence tagging has, in general, low EO and low CT score; i.e., labels are not edited much and a final decision is reached early. That does not hold for~BERT, whose CT score and EO is, in general, higher. CT score and EO in sequence classification are also higher because the label in this case should capture a more global representation, which cannot reasonably be expected to be very good when only a small part of the sequence has yet been seen.

When it comes to RC (correctness relative to the final output), again~BERT has worse results than other models, especially for tagging. For sequence classification, BERT's performance is more in line with the other models. Achieving high RC is desirable because it means that, most of the time, the partial outputs are correct prefixes of the non-incremental output and can be trusted, at least to the same degree that the final result can be trusted. 

\begin{figure*}[h]
	\centering
	\begin{subfigure}[t]{0.49\textwidth}
		\centering
		\includegraphics[trim={0cm 0cm 0cm 0cm}, clip, width=7.9cm]{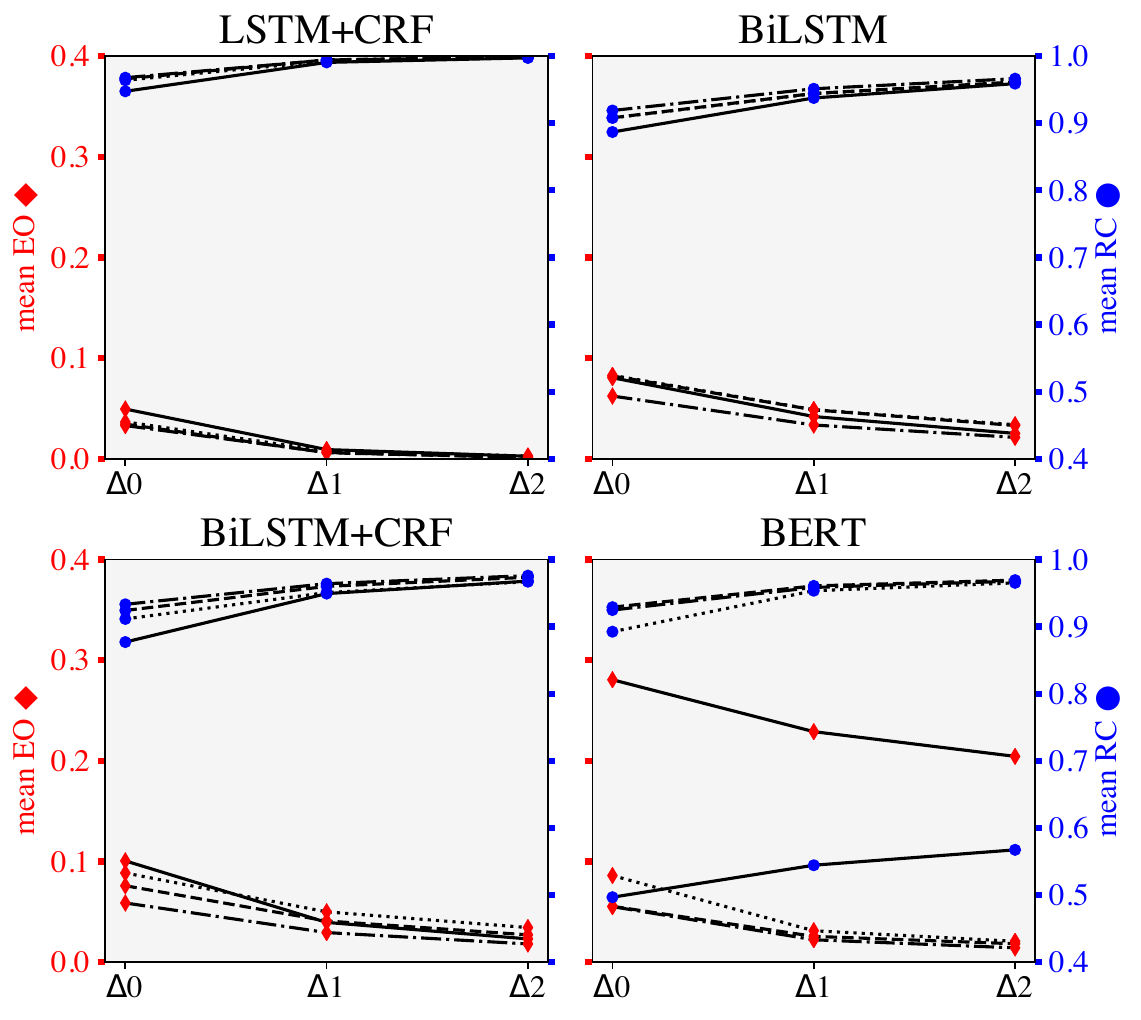} 
		\caption{Sequence tagging}
		\label{fig:strategies-tagging}
	\end{subfigure}
	\centering
	\begin{subfigure}[t]{0.49\textwidth}
		\centering
		\includegraphics[trim={0cm 0cm 0cm 0cm},clip, width=7.9cm]{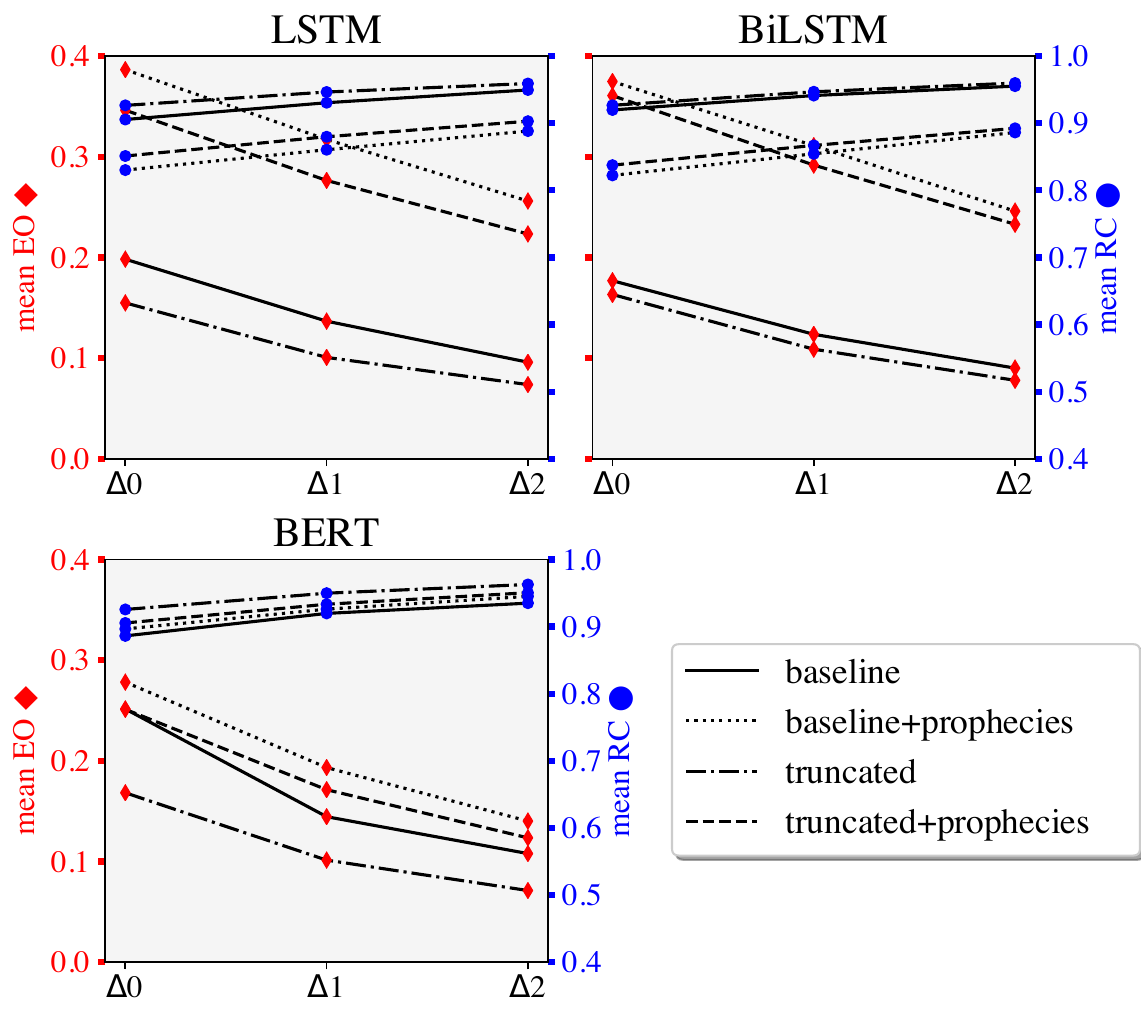} 
		\caption{Sequence classification}
		\label{fig:strategies-classification}
	\end{subfigure}
	
	\caption{Comparison of mean Edit Overhead and mean Relative-Correctness on the baseline incremental interface and the three strategies using observations from all tasks except SRL.} 
	\label{fig:aggregate-comparison}
\end{figure*}

This overview shows that although~BERT's non-incremental performance is normally the highest, the quality of its incremental outputs is more unstable. The next step is examining the effect of the three strategies that seek to improve the quality and stability of incremental outputs. Figure~\ref{fig:aggregate-comparison} shows that truncated training is always beneficial, as is delayed evaluation, with both strategies reducing EO and increasing RC. The fact that delay helps in all cases indicates that most substitutions happen in the last or last but one label (the right frontier, given the current prefix), or, in other words, that even having a limited right context improves quality substantially. \footnote{
	Results for SRL are not included in Figure \ref{fig:aggregate-comparison}, because they go in the opposite direction of all other tasks. Since this task depends on the predicate embedding, both truncating the training sequence or adding a right context with no predicate information reduces performance in most cases, except for~BERT. See Appendix for results separated by model and task.}

Prophecies are detrimental in classification tasks, but they help in some tagging tasks, especially for~BERT. Most importantly, any of the strategies cause a great improvement to~BERT's incremental performance in sequence tagging, making its metrics be on the same level as other models while retaining its superior non-incremental quality.

\begin{figure}
	\centering
	\begin{subfigure}[t]{\linewidth}
		\centering
		\includegraphics[width=6cm]{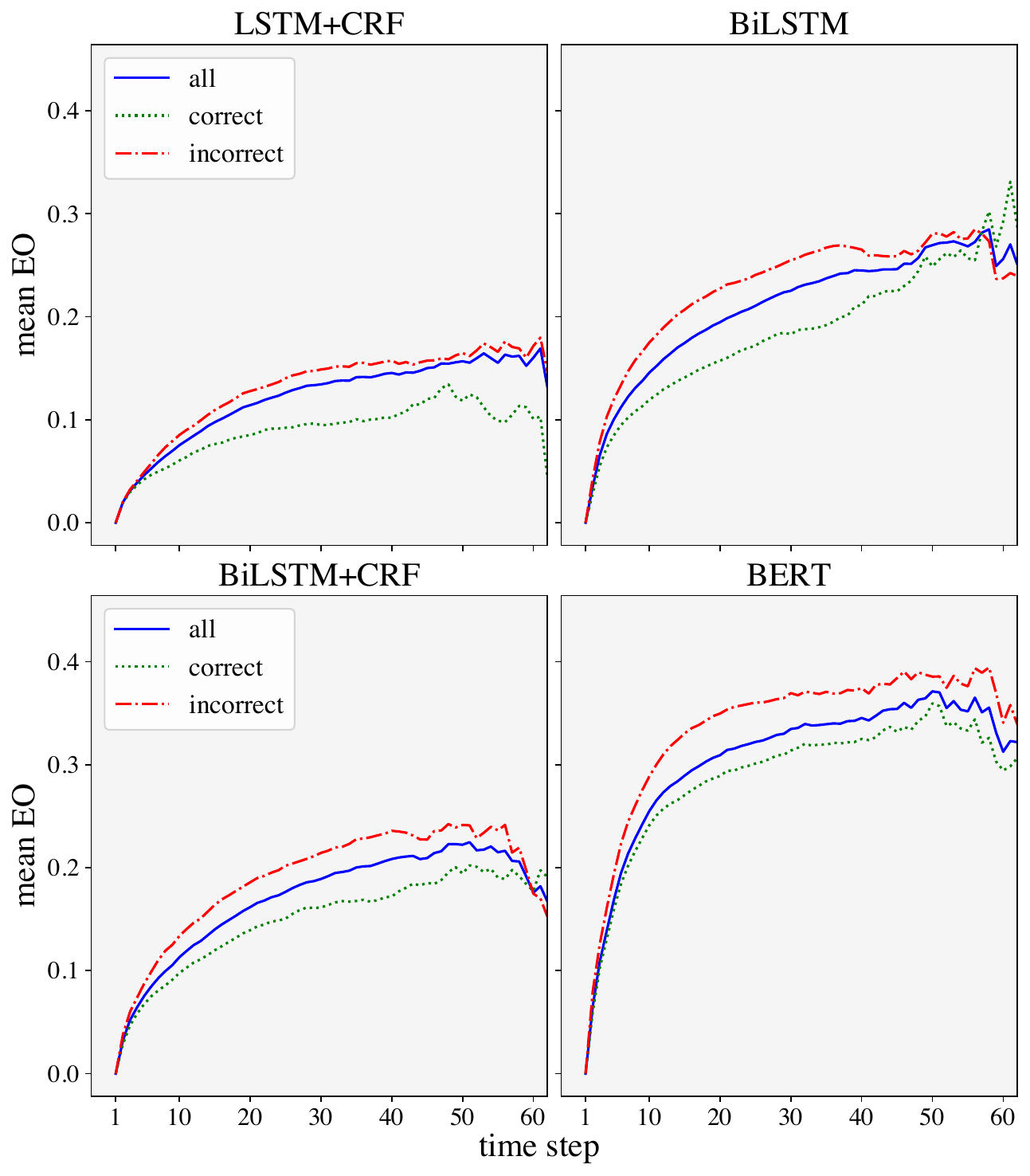}
		\caption{Sequence tagging}
		\label{fig:eo-time-tagging}
	\end{subfigure}
	\centering
	\begin{subfigure}[t]{\linewidth}
		\centering
		\includegraphics[width=8cm]{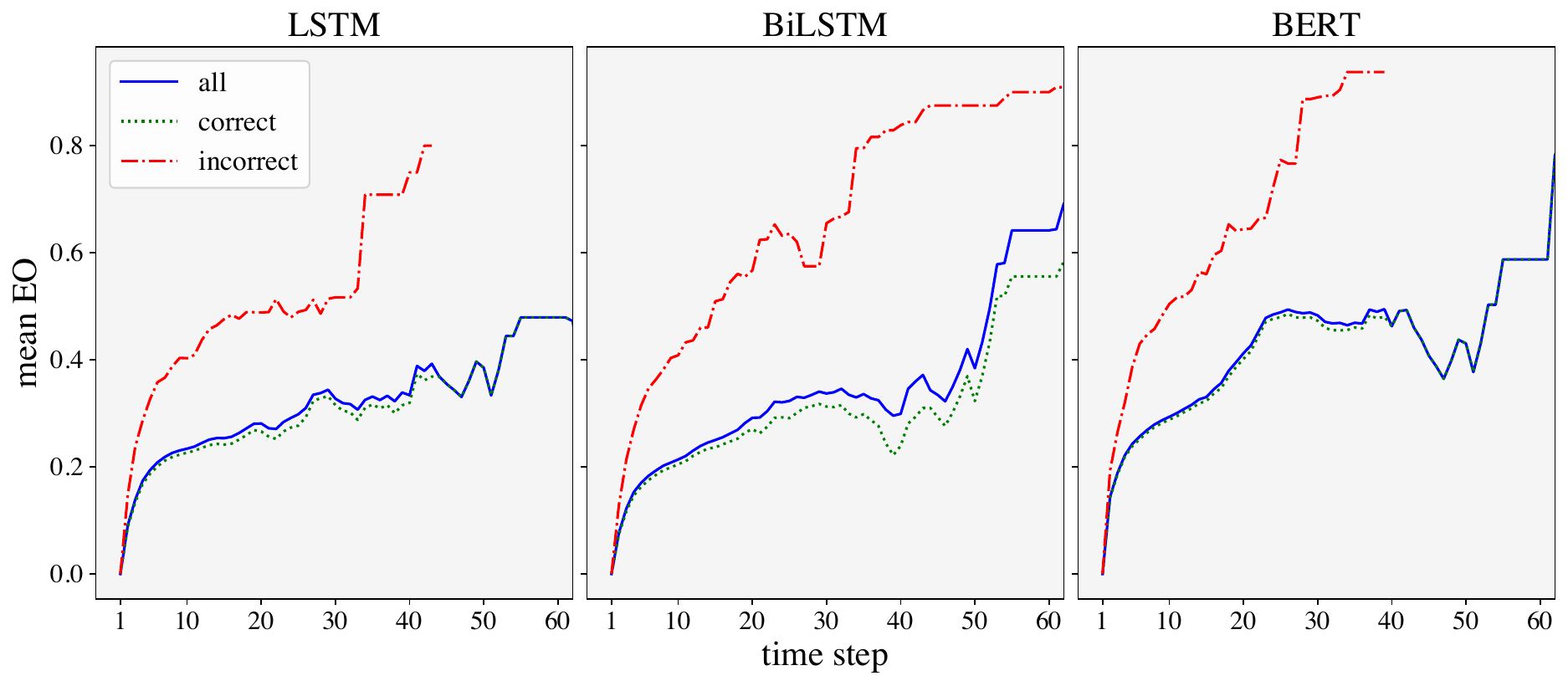} 
		\caption{Sequence classification}
		\label{fig:eo-time-classification}
	\end{subfigure}
	
	\caption{Development of mean Edit Overhead over time using observations from all tasks. \textit{correct} means that all final output labels of a sentence are right and \textit{incorrect} means that at least one label of the final output sequence is wrong. All models are more unstable when their non-incremental final output is incorrect with respect to the gold output.} 
	\label{fig:eo-overtime}
\end{figure}

Note that while CT and RC can only be measured once the final output is available, an estimate of EO may be evaluated on the fly if we consider the edits and additions up to the last output. Figure \ref{fig:eo-overtime} shows how the mean EO evolves, breaking out the results for cases where the non-incremental final output will be correct and those where it will not with respect to the gold labels. We can observe an intriguing pattern: the mean EO grows faster for cases where the final response will be wrong; this is most pronounced for the sequence classification task. It might be possible to use this observation as an indication of how much to trust the final result: If the incremental computation was more unstable than the average, we should not expect the final result to be good. However, initial experiments on building a classifier based on the instability of partial outputs have so far not been successful in cashing in on that observation.

\section{Discussion and conclusion}
\label{conclusion}

We show that bidirectional encoders can be adapted to work under an incremental interface without a too drastic impact on their performance.
Even though the training (being done on complete sequences) differs from the testing situation (which exposes the model to partial input), the incremental metrics of most models are, in general, good: in sequence tagging, edit overhead is low, final decisions are taken early, and often partial outputs are a correct prefix of the complete non-incremental output. Sequence classification is more unstable because, at initial steps, there is a higher level of uncertainty on what is coming next.
Our experiments show that the deficiencies of~BERT in the incremental metrics can be mitigated with some adaptations (truncated training or prophecies together with delay), which make its incremental quality become as good as those of other models.

Since the semantic information is only kept encoded for a few steps in RNNs~\citep{hupkes2018}, this may be a reason why delay causes incremental metrics to be much better. If long-range dependencies are not captured, only neighboring words exert more influence in the choice of a label, so after seeing two words in the right context, the system rarely revises labels further back.
BERT, having access to the whole sentence at any time step, is less stable because new input can cause it to reassess past labels more easily.

Besides, we also found evidence of different behavior of the instability of partial outputs between correct and incorrect output sequences, which could potentially be a signal of later lower quality. This could be used, for example, in dialog systems: if edit overhead gets too high, a clarification request should be made. A follow-up idea is training a classifier that predicts more precisely how likely it is that the final labels will be accurate based on the development of EO. However, our initial experiments on building such classifier were not successful. We suppose this is due to the fact that, in our datasets, incorrect final output sequences still usually have more than 90\% correct labels, so the learnable signal may be too weak.

The use of~GPT-2 prophecies led to promising improvements for~BERT in sequence tagging. We see room for improvement, \textit{e.g.} resorting to domain adaptation to make prophecies be more related to each genre. A natural extension is training a language model that generates the prophecies together with the encoder.

Finally, we believe that using attention mechanisms to study the grounding of the edits, similarly to the ideas in~\citet{kohn2018incremental}, can be an important step towards understanding how the preliminary representations are built and decoded; we want to test this as well in future work.

\section*{Acknowledgments}

We are thankful to the three anonymous reviewers and to our colleagues in the Foundations of Computational Linguistics Lab at the University of Potsdam for their valuable comments and insights. D.S. acknowledges the support by the Deutsche Forschungsgemeinschaft (DFG), grant SCHL 845/9-1.

\section*{Updates and Corrections}
We have fixed a wrong step in the implementation of the delayed EO and updated Figure 5, Figure 8 and Table 10 accordingly. The text remains the same since no conclusions changed. A few citations were missing that we will add here: The GPT-2 model was proposed by \citet{radford2019language}. The incremental interface paradigm was also defined by \citet{schlangen2011general} as \textit{restart-incrementality}. We use the preprocessed data and splits for SNIPS and ATIS made available by \citet{e-etal-2019-novel} and relied on \url{https://github.com/yuchenlin/OntoNotes-5.0-NER-BIO/} and \url{https://github.com/ontonotes/conll-formatted-ontonotes-5.0} to preprocess OntoNotes.

\bibliographystyle{acl_natbib}
\bibliography{emnlp2020}


\clearpage
\appendix

\section{Reproducibility}

Here we describe more details about hyperparameters and the models. 
Table \ref{table:hyp-search} to Table \ref{table:rc-delay} present more information about the hyperparameter search and the datasets as well as explicit results for each model and task, to be used for reproducibility purposes.\\

\textbf{Data}

\begin{itemize}
	\item We use only the~WSJ section of OntoNotes, with the train-valid-test splits provided by~\citet{pradhan2013ontonotes}, as well as their conversion to~CoNLL format. 
	\item The PosNeg and ProsCons datasets, which are not published with a train-valid-test split, were divided into 70\%-10\%-20\% sets randomly.
 \item We removed the two longest sentences ($>$200 words) in Pros/Cons the dataset because they were infeasible to compute with~BERT.
 \item Sentences longer than 60 words were removed in the hyperparameter search phase only.
\end{itemize}

\textbf{Implementation}
\begin{itemize}
 \item Hyperparameter search is done for each dataset in both the LSTM model and BERT.
 \item We use Comet's Bayes algorithm\footnote{\url {https://www.comet.ml/docs/python-sdk/introduction-optimizer/} }, which balances exploration and exploitation and, in our experiments, tries to maximize the accuracy (for sequence classification) of F1 score (for sequence tagging) in the validation set. 
 \item We set the maximum number of iterations to 50, but early stopping happens if no improvement is seen during 10 iterations.
\item The best configuration of the LSTM model is also used for the LSTM+CRF, BiLSTM, and BiLSTM+CRF models.
\item We use a five-dimensional embedding for the binary predicates in the SRL task.
\item Hidden states are initialized as 0. 
\item All the weights and biases are initialized with PyTorch's default (uniformly sampled from $(-\sqrt{hidden\_size}), \sqrt{hidden\_size}$). 
\item Dropout is implemented after the embedding layer and after the encoder layer with the same value.
\item PyTorch's and Numpy's manual seeds are set to 2204 for all experiments.
\item All experiments were run on a GPU GeForce GTX 1080 Ti.

\end{itemize}

\vspace{5cm}

\begin{table}[h]
	\small 
	\begin{center}
		\begin{tabular}{l l}
			\toprule
			\textbf{Hyperparameter}& \textbf{LSTM} \\
			\cmidrule{1-2} 
			Batch size & 32, 64, 128, 512 \\
			Clipping & 0.25, 0.5, 1 \\
			Dropout & 0.1, 0.2, 0.3, 0.5  \\
			Embedding dimension & 50, 100, 200, 300 \\
			Hidden layer dimension & 50, 100, 150, 200, 300 \\
			Learning rate & 0.1, 0.01, 0.001 \\
			Number of layers & 1, 2, 3, 4 \\
			\cmidrule{2-2}
			& \textbf{BERT} \\
			\cmidrule{2-2}
			Batch size & 16, 32 \\
			Clipping & 0.25, 0.5, 1 \\
			Learning rate & 5e-05, 3e-05, 2e-05, 1e-05 \\
			
			\bottomrule
		\end{tabular}
	\end{center}
	\caption{Hyperparameter search space.}
	\label{table:hyp-search}
\end{table}

\begin{table*}[h]
	\small
	\begin{center}
		
		\begin{tabular}{l p{1cm} p{1.1cm} p{1cm} p{1cm} p{1cm} p{1.2cm} p{1cm} p{1cm} p{1.2cm}} 
			\toprule 
			& & & \multicolumn{7}{c}{\bf best configuration}  \\ 
			\cmidrule{4-10} 
			\bf Task/Model  & \bf search trials & \bf avrg. runtime &  batch size & clipping & dropout & embedding layer & hidden layer & learning rate & number of layers  \\ 
			\midrule
			\sc Sequence tagging & & & & & & & & &  \\ 
			\cmidrule{1-1} 
			\hspace{0.2cm}Chunk & 24 & 6 & 32 & 0.5 & 0.5 & 300 & 200 & 0.001 & 1 \\ 
			\hspace{0.2cm}Named Entity Recognition & 24 & 16 & 32 & 1 & 0.3 & 50 & 300 & 0.001 & 2 \\ 
			\hspace{0.2cm}Part-of-Speech Tagging & 26 & 19 & 32 & 0.25 & 0.5 & 300 & 150	& 0.001 & 3 \\ 
			\hspace{0.2cm}Semantic Role Labeling & 34 & 33 & 32 & 1 & 0.3 & 100 & 300 & 0.001 & 2 \\ 
			\hspace{0.2cm}Slot Filling (ATIS) & 23 & 2 & 32 & 0.25 & 0.5 & 50 & 200	 & 0.01 & 1 \\ 
			\hspace{0.2cm}Slot Filling (SNIPS) & 39 & 5 & 32 & 1 & 0.5 & 50 & 150 & 0.001 & 2 \\ 
			& & & & & & & & & \\
			\sc Sequence classification & & & & & & & & & \\ 
			\cmidrule{1-1} 
			\hspace{0.2cm}Intent (ATIS) & 60 & 1 & 64 & 0.5 & 0.1 & 200 & 200 & 0.001 & 2 \\ 
			\hspace{0.2cm}Intent (SNIPS) & 66 & 3 & 64 & 0.5 & 0.5 & 100 & 300 & 0.001 & 1 \\ 
			\hspace{0.2cm}Positive/Negative & 27 & 1 & 64 & 0.5	& 0.5 & 300 & 100 & 0.01 & 3 \\ 
			\hspace{0.2cm}Pros/Cons & 38 & 6 & 32 & 0.5 & 0.5 & 300 & 150 & 0.001 & 4 \\ 
			
			\bottomrule 
		\end{tabular} 
		
	\end{center}
	\caption{Hyperparameter search for LSTM model. The best configuration was also used for LSTM+CRF, BiLSTM and BiLSTM+CRF. Runtime in minutes.}
	\label{table:reproducibility-hyperparams}
\end{table*}

\begin{table*}[h]
	\small
	\begin{center}
		
		\begin{tabular}{l p{1cm} p{1.1cm} p{1cm} p{1cm} p{1cm}} 
			\toprule 
			& &  & \multicolumn{3}{c}{\bf best configuration}  \\ 
			\cmidrule{4-6} 
			\bf Task/Model  & \bf search trials & \bf avrg. runtime &  batch size & clipping & learning rate  \\ 
			\midrule
			\sc Sequence tagging & & & & &  \\ 
			\cmidrule{1-1} 
			\hspace{0.2cm}Chunk & 8 & 24 & 16 & 0.25	& 2e-05 \\
			\hspace{0.2cm}Named Entity Recognition & 9 & 99 & 16	& 1	& 2e-05 \\
			\hspace{0.2cm}Part-of-Speech Tagging & 7 & 62 & 16 & 0.5 & 3e-05  \\
			\hspace{0.2cm}Semantic Role Labeling & 9 & 471 & 16 & 0.5 & 2e-05\\
			\hspace{0.2cm}Slot Filling (ATIS) & 10 & 15 & 32 & 0.25 & 5e-05 \\ 
			\hspace{0.2cm}Slot Filling (SNIPS) & 11 & 33 & 16 & 0.5 & 2e-05\\
			& & & & & \\
			\sc Sequence classification & & & & & \\ 
			\cmidrule{1-1} 
			\hspace{0.2cm}Intent (ATIS) & 10 & 12 & 32 & 0.25 & 5e-05 \\
			\hspace{0.2cm}Intent (SNIPS) & 10 & 23 & 16 & 0.5 & 3e-05 \\
			\hspace{0.2cm}Positive/Negative & 10 &  4 & 16 & 0.25 & 2e-05\\ 
			\hspace{0.2cm}Pros/Cons & 10 & 70 & 32 & 0.5 & 3e-05 \\
			
			\bottomrule 
		\end{tabular} 
		
	\end{center}
	\caption{Hyperparameter search for BERT model. Runtime in minutes.}
	\label{table:reproducibility-hyperparams-bert}
\end{table*}

\begin{table*}[h]
	\small
	\begin{center}
		
		\begin{tabular}{l r r r r r} 
			\toprule 
			& \multicolumn{5}{c}{\bf Model}  \\ 
			\cmidrule{2-6} 
			\bf Task &  LSTM & LSTM+CRF &  BiLSTM & BiLSTM+CRF & BERT  \\ 
			\midrule

			Chunking & 5,775,023 & 5,775,598 & 6,181,223 & 6,181,798 & 108,327,959\\
			NER & 2,937,587 & 2,939,030 & 4,813,487 & 4,814,930 & 108,338,725\\
			POS & 11,330,748 & 11,333,148 & 12,331,548 & 12,333,948 & 108,347,184\\
			SRL & 4,829,016 & 4,840,464 & 6,791,616 & 6,803,064 & 108,391,786\\
			SlotATIS & 270,327 & 286,710 & 497,327 & 513,710 & 108,407,935\\
			SlotSNIPS & 876,522 & 881,850 & 1,369,722 & 1,375,050 & 108,365,640\\
			IntentATIS & 821,226 & - & 1,789,626 & - & 108,330,266\\
			IntentSNIPS & 1,611,007 & - & 2,095,507 & - & 108,315,655\\
			PosNeg & 1,764,402 & - & 2,247,002 & - & 108,311,810\\
			ProsCons & 4,905,302 & - & 6,260,402 & - & 108,311,810\\
			
			\bottomrule 
		\end{tabular} 
		
	\end{center}
	\caption{Number of parameters in each model.}
	\label{table:reproducibility-num-params}
\end{table*}


\begin{table*}[h]
	\small
	\begin{center}
		\begin{tabular}{l l l r r r r}
			\toprule
			\bf Task & \bf Dataset & \bf Reference & \bf Labels & \bf Train & \bf Valid & \bf Test   \\ 
			\midrule
			\sc Sentence tagging & & & & & & \\
			\hspace{0.1cm} Chunking & CoNLL 2000 & ~\citet{conllchunk} & 23 & 7,922 & 1,014 & 2,012 \\
			\hspace{0.1cm} Named Entity Recognition & OntoNotes 5.0 & ~\citet{ontonotes} & 37 & 30,060 & 5,315 & 1,640	 \\
			\hspace{0.1cm} Part-of-Speech Tagging & OntoNotes 5.0 & ~\citet{ontonotes} & 48 & 30,060 & 5,315 & 1,640\\
			\hspace{0.1cm} Semantic Role Labeling & OntoNotes 5.0 & ~\citet{ontonotes} & 106 & 83,920 & 15,208 & 4,781\\
			\hspace{0.1cm} Slot Filling & ATIS &  ~\citet{hemphill1990atis} & 127 & 4,478 & 500 & 893 \\
			\hspace{0.1cm} Slot Filling & SNIPS & ~\citet{coucke2018snips} & 72 & 13,084 & 700 & 700   \\
			\sc Sentence classification &  & & & & & \\ 
			\hspace{0.1cm} Intent & ATIS & ~\citet{hemphill1990atis} & 26 & 4,478 & 500 & 893\\
			\hspace{0.1cm} Intent & SNIPS & ~\citet{coucke2018snips} & 7 & 13,084 & 700 & 700 \\
			\hspace{0.1cm} Sentiment & Positive/Negative & ~\citet{kotzias2015posneg} & 2 & 2,100 & 300 & 600 \\
			\hspace{0.1cm} Sentiment &Pros/Cons &  ~\citet{ganapathibhotla2008proscons} & 2 & 32,088 & 4,602 & 9,175\\
			\bottomrule
		\end{tabular}
	\end{center}
	\caption{Tasks and datasets.}
	\label{table:data}
\end{table*}

\begin{table*}[!h]
	\small
	\centering
\begin{tabular}{r c c c c c c} 
	\toprule 
	& & \multicolumn{5}{c}{\bf Model } \\ 
	\cmidrule{2-7} 
	\bf Task & \bf Metric & \bf LSTM & \bf LSTM+CRF & \bf BiLSTM & \bf BiLSTM+CRF & \bf BERT \\ 
	\cmidrule{2-7} 
	Chunk & & 88.39 (88.42) & 91.52 (90.79) & 91.67 (90.93) & 92.53 (91.76) & 97.53 (97.00)\\ 
	Named Entity Recognition & & 68.60 (67.36) & 85.22 (82.08) & 86.77 (83.02) & 87.86 (84.78) & 92.05 (89.38)\\ 
	Semantic Role Labeling & F1 Score & 52.55 (49.78) & 71.48 (67.22) & 77.53 (70.57) & 84.16 (77.91) & 89.29 (82.81)\\ 
	Slot Filling (ATIS) & (\%) & 95.76 (94.54) & 96.93 (95.91) & 97.16 (96.07) & 97.33 (96.62) & 98.39 (96.67)\\ 
	Slot Filling (SNIPS) & & 82.86 (80.12) & 90.36 (86.12) & 90.47 (84.90) & 91.60 (87.26) & 95.56 (88.52)\\ 
	\cmidrule{2-7} 
	Intent (ATIS) & & 98.40 (90.60) & -  & 98.20 (90.40) & -  & 98.80 (93.60) \\ 
	Intent (SNIPS) & & 99.71 (95.14) & -  & 99.57 (94.00) & -  & 99.29 (93.57) \\ 
	Part-of-Speech Tagging & Accuracy & 94.72 (94.00)  & 95.75 (94.88)  & 96.24 (95.54)  & 96.27 (95.54)  & 97.90 (97.45) \\ 
	Positive/Negative & (\%) & 85.33 (70.67) & -  & 85.67 (70.67) & -  & 95.67 (76.33) \\ 
	Pros/Cons & & 94.59 (90.24) & -  & 94.74 (90.48) & -  & 96.02 (91.81) \\ 
	\bottomrule 
\end{tabular} 
	
	\caption{Non-incremental performance all models on validation sets for the purpose of reproducibility. Values in parentheses refer to using truncated samples during training.}
	\label{table:valid-performance}
\end{table*}

\begin{table*}[!h]
	\small
	\centering
	
	\begin{tabular}{r c c c c c} 
		\toprule 
		& \multicolumn{5}{c}{\bf Model } \\ 
		\cmidrule{2-6} 
		\bf Task & \bf LSTM & \bf LSTM+CRF & \bf BiLSTM & \bf BiLSTM+CRF & \bf BERT \\ 
		\cmidrule{2-6} 
		& \multicolumn{5}{c}{Sentence level correctness (\%)}  \\ 
		\cmidrule{2-6} 
		\sc Sequence tagging & & & & &  \\ 
		\cmidrule{1-1} 
		Chunk & 33.95 (28.03) & 44.04 (37.48) & 43.74 (36.38) & 48.56 (43.39) & 71.42 (69.68)\\ 
		Named Entity Recognition & 51.04 (48.60) & 70.61 (66.83) & 74.02 (68.05) & 75.00 (72.44) & 85.61 (83.17)\\ 
		Part-of-Speech Tagging & 36.77 (34.02) & 45.12 (42.68) & 49.33 (47.38) & 50.55 (49.21) & 63.41 (60.49)\\ 
		Semantic Role Labeling & 6.82 (6.65) & 24.14 (21.36) & 48.90 (41.92) & 56.14 (49.42) & 67.85 (63.75)\\ 
		Slot Filling (ATIS) & 84.10 (75.92) & 87.46 (78.72) & 86.34 (76.48) & 87.68 (79.06) & 89.36 (84.88)\\ 
		Slot Filling (SNIPS) & 61.29 (54.71) & 75.00 (66.86) & 79.00 (69.14) & 81.71 (71.86) & 89.29 (84.43)\\ 
		& & & & & \\ 
		\sc Sequence classification & & & & &  \\ 
		\cmidrule{1-1} 
		Intent (ATIS) & 96.86 (93.06)& -  & 95.74 (93.62)& -  & 97.31 (95.86)\\ 
		Intent (SNIPS) & 96.86 (97.43)& -  & 97.43 (97.43)& -  & 97.57 (97.71)\\ 
		Positive/Negative & 82.17 (72.83)& -  & 83.33 (75.67)& -  & 93.83 (92.50)\\ 
		Pros/Cons & 94.51 (93.85)& -  & 94.40 (93.65)& -  & 95.74 (95.17)\\ 
		\bottomrule 
	\end{tabular} 
	
	\caption{Sentence-level non-incremental performance of all models on test sets (the same as accuracy in sequence classification). Values in parentheses refer to using truncated samples during training.}
	\label{table:performance-sentence-level}
\end{table*}


\begin{table*}[h!]
	\small
	\begin{center}
		
		\begin{tabular}{l c c c c c c } 
			\toprule 
			\multirow{2}{*}{\bf Task/Model} & \multicolumn{3}{c}{\bf Metrics } & \multicolumn{3}{c}{\bf Metrics with prophecies} \\ 
			\cmidrule{2-4} \cmidrule{5-7}
			& \bf EO & \bf CT & \bf RC & \bf EO & \bf CT & \bf RC \\ 
			\midrule
			\sc Sequence tagging & & & & & & \\ 
			\cmidrule{1-1} 
			\hspace{0.2cm}Chunk & & & & & & \\ 
			\hspace{0.4cm} \sc LSTM& 0.00 (0.00) & 0.00 (0.00) & 1.00 (1.00)& 0.00 (0.00) & 0.00 (0.00) & 1.00 (1.00)\\ 
			\hspace{0.4cm} \sc LSTM+CRF& 0.06 (0.03) & 0.01 (0.00) & 0.94 (0.97)& 0.03 (0.03) & 0.00 (0.00) & 0.97 (0.97)\\ 
			\hspace{0.4cm} \sc BiLSTM& 0.10 (0.08) & 0.03 (0.02) & 0.84 (0.89)& 0.09 (0.11) & 0.02 (0.03) & 0.90 (0.87)\\ 
			\hspace{0.4cm} \sc BiLSTM+CRF& 0.16 (0.07) & 0.04 (0.02) & 0.79 (0.92)& 0.10 (0.09) & 0.02 (0.02) & 0.91 (0.91)\\ 
			\hspace{0.4cm} \sc BERT& 0.17 (0.06) & 0.06 (0.02) & 0.67 (0.91)& 0.10 (0.06) & 0.02 (0.01) & 0.87 (0.92)\\ 
			\hspace{0.2cm}Named Entity Recognition & & & & & & \\ 
			\hspace{0.4cm} \sc LSTM& 0.00 (0.00) & 0.00 (0.00) & 1.00 (1.00)& 0.00 (0.00) & 0.00 (0.00) & 1.00 (1.00)\\ 
			\hspace{0.4cm} \sc LSTM+CRF& 0.05 (0.04) & 0.01 (0.01) & 0.95 (0.96)& 0.04 (0.04) & 0.01 (0.01) & 0.96 (0.96)\\ 
			\hspace{0.4cm} \sc BiLSTM& 0.07 (0.06) & 0.02 (0.01) & 0.91 (0.93)& 0.08 (0.08) & 0.02 (0.02) & 0.91 (0.92)\\ 
			\hspace{0.4cm} \sc BiLSTM+CRF& 0.08 (0.06) & 0.02 (0.01) & 0.92 (0.94)& 0.09 (0.08) & 0.02 (0.02) & 0.92 (0.93)\\ 
			\hspace{0.4cm} \sc BERT& 0.17 (0.06) & 0.12 (0.02) & 0.49 (0.93)& 0.08 (0.06) & 0.02 (0.01) & 0.90 (0.93)\\ 
			\hspace{0.2cm}Part-of-Speech Tagging & & & & & & \\ 
			\hspace{0.4cm} \sc LSTM& 0.00 (0.00) & 0.00 (0.00) & 1.00 (1.00)& 0.00 (0.00) & 0.00 (0.00) & 1.00 (1.00)\\ 
			\hspace{0.4cm} \sc LSTM+CRF& 0.05 (0.03) & 0.01 (0.01) & 0.95 (0.96)& 0.03 (0.03) & 0.01 (0.01) & 0.97 (0.97)\\ 
			\hspace{0.4cm} \sc BiLSTM& 0.08 (0.06) & 0.02 (0.01) & 0.89 (0.93)& 0.07 (0.06) & 0.02 (0.02) & 0.92 (0.93)\\ 
			\hspace{0.4cm} \sc BiLSTM+CRF& 0.09 (0.06) & 0.02 (0.01) & 0.89 (0.92)& 0.07 (0.06) & 0.02 (0.02) & 0.92 (0.93)\\ 
			\hspace{0.4cm} \sc BERT& 0.52 (0.05) & 0.54 (0.01) & 0.30 (0.93)& 0.07 (0.05) & 0.02 (0.01) & 0.90 (0.93)\\ 
			\hspace{0.2cm}Semantic Role Labeling & & & & & & \\ 
			\hspace{0.4cm} \sc LSTM& 0.00 (0.00) & 0.00 (0.00) & 1.00 (1.00)& 0.00 (0.00) & 0.00 (0.00) & 1.00 (1.00)\\ 
			\hspace{0.4cm} \sc LSTM+CRF& 0.21 (0.29) & 0.07 (0.09) & 0.71 (0.83)& 0.32 (0.26) & 0.10 (0.09) & 0.82 (0.85)\\ 
			\hspace{0.4cm} \sc BiLSTM& 0.31 (0.37) & 0.15 (0.16) & 0.59 (0.60)& 0.34 (0.40) & 0.16 (0.19) & 0.62 (0.62)\\ 
			\hspace{0.4cm} \sc BiLSTM+CRF& 0.28 (0.31) & 0.15 (0.15) & 0.65 (0.72)& 0.28 (0.35) & 0.15 (0.17) & 0.71 (0.72)\\ 
			\hspace{0.4cm} \sc BERT& 0.43 (0.33) & 0.31 (0.14) & 0.25 (0.70)& 0.22 (0.26) & 0.12 (0.14) & 0.69 (0.67)\\ 
			\hspace{0.2cm}Slot Filling (ATIS) & & & & & & \\ 
			\hspace{0.4cm} \sc LSTM& 0.00 (0.00) & 0.00 (0.00) & 1.00 (1.00)& 0.00 (0.00) & 0.00 (0.00) & 1.00 (1.00)\\ 
			\hspace{0.4cm} \sc LSTM+CRF& 0.02 (0.01) & 0.01 (0.00) & 0.98 (0.98)& 0.01 (0.03) & 0.00 (0.01) & 0.99 (0.97)\\ 
			\hspace{0.4cm} \sc BiLSTM& 0.02 (0.02) & 0.01 (0.01) & 0.97 (0.98)& 0.02 (0.02) & 0.01 (0.01) & 0.97 (0.98)\\ 
			\hspace{0.4cm} \sc BiLSTM+CRF& 0.03 (0.01) & 0.01 (0.01) & 0.97 (0.98)& 0.04 (0.02) & 0.01 (0.01) & 0.95 (0.98)\\ 
			\hspace{0.4cm} \sc BERT& 0.22 (0.03) & 0.19 (0.01) & 0.56 (0.97)& 0.06 (0.03) & 0.02 (0.01) & 0.93 (0.97)\\ 
			\hspace{0.2cm}Slot Filling (SNIPS) & & & & & & \\ 
			\hspace{0.4cm} \sc LSTM& 0.00 (0.00) & 0.00 (0.00) & 1.00 (1.00)& 0.00 (0.00) & 0.00 (0.00) & 1.00 (1.00)\\ 
			\hspace{0.4cm} \sc LSTM+CRF& 0.06 (0.04) & 0.03 (0.02) & 0.93 (0.96)& 0.10 (0.07) & 0.05 (0.03) & 0.90 (0.93)\\ 
			\hspace{0.4cm} \sc BiLSTM& 0.12 (0.08) & 0.07 (0.05) & 0.84 (0.90)& 0.17 (0.13) & 0.10 (0.07) & 0.81 (0.85)\\ 
			\hspace{0.4cm} \sc BiLSTM+CRF& 0.11 (0.08) & 0.06 (0.04) & 0.88 (0.91)& 0.17 (0.14) & 0.10 (0.07) & 0.82 (0.86)\\ 
			\hspace{0.4cm} \sc BERT& 0.37 (0.08) & 0.38 (0.04) & 0.41 (0.91)& 0.12 (0.09) & 0.06 (0.05) & 0.86 (0.90)\\ 
			& & & & & & \\ 
			\sc Sequence classification & & & & & & \\ 
			\cmidrule{1-1} 
			\hspace{0.2cm}Intent (ATIS) & & & & & & \\ 
			\hspace{0.4cm} \sc LSTM& 0.48 (0.21) & 0.27 (0.12) & 0.77 (0.91)& 0.77 (0.78) & 0.72 (0.68) & 0.52 (0.54)\\ 
			\hspace{0.4cm} \sc LSTM+CRF& - & - & -& - & - & - \\ 
			\hspace{0.4cm} \sc BiLSTM& 0.40 (0.24) & 0.23 (0.14) & 0.85 (0.90)& 0.66 (0.77) & 0.38 (0.73) & 0.70 (0.50)\\ 
			\hspace{0.4cm} \sc BiLSTM+CRF& - & - & -& - & - & - \\ 
			\hspace{0.4cm} \sc BERT& 0.49 (0.20) & 0.20 (0.13) & 0.84 (0.91)& 0.38 (0.29) & 0.19 (0.16) & 0.88 (0.90)\\ 
			\hspace{0.2cm}Intent (SNIPS) & & & & & & \\ 
			\hspace{0.4cm} \sc LSTM& 0.31 (0.24) & 0.20 (0.14) & 0.85 (0.90)& 0.57 (0.49) & 0.48 (0.39) & 0.71 (0.77)\\ 
			\hspace{0.4cm} \sc LSTM+CRF& - & - & -& - & - & - \\ 
			\hspace{0.4cm} \sc BiLSTM& 0.26 (0.23) & 0.19 (0.13) & 0.86 (0.91)& 0.55 (0.49) & 0.47 (0.41) & 0.71 (0.76)\\ 
			\hspace{0.4cm} \sc BiLSTM+CRF& - & - & -& - & - & - \\ 
			\hspace{0.4cm} \sc BERT& 0.30 (0.22) & 0.21 (0.13) & 0.83 (0.91)& 0.40 (0.35) & 0.26 (0.20) & 0.83 (0.86)\\ 
			\hspace{0.2cm}Positive/Negative & & & & & & \\ 
			\hspace{0.4cm} \sc LSTM& 0.38 (0.40) & 0.31 (0.31) & 0.78 (0.79)& 0.65 (0.68) & 0.59 (0.62) & 0.74 (0.72)\\ 
			\hspace{0.4cm} \sc LSTM+CRF& - & - & -& - & - & - \\ 
			\hspace{0.4cm} \sc BiLSTM& 0.37 (0.39) & 0.29 (0.30) & 0.82 (0.83)& 0.63 (0.58) & 0.51 (0.49) & 0.76 (0.77)\\ 
			\hspace{0.4cm} \sc BiLSTM+CRF& - & - & -& - & - & - \\ 
			\hspace{0.4cm} \sc BERT& 0.58 (0.45) & 0.56 (0.30) & 0.64 (0.80)& 0.56 (0.55) & 0.43 (0.43) & 0.79 (0.79)\\ 
			\hspace{0.2cm}Pros/Cons & & & & & & \\ 
			\hspace{0.4cm} \sc LSTM& 0.15 (0.13) & 0.11 (0.09) & 0.93 (0.94)& 0.32 (0.27) & 0.26 (0.21) & 0.88 (0.90)\\ 
			\hspace{0.4cm} \sc LSTM+CRF& - & - & -& - & - & - \\ 
			\hspace{0.4cm} \sc BiLSTM& 0.14 (0.14) & 0.10 (0.10) & 0.94 (0.94)& 0.32 (0.30) & 0.28 (0.24) & 0.85 (0.88)\\ 
			\hspace{0.4cm} \sc BiLSTM+CRF& - & - & -& - & - & - \\ 
			\hspace{0.4cm} \sc BERT& 0.20 (0.14) & 0.15 (0.10) & 0.91 (0.94)& 0.24 (0.22) & 0.17 (0.16) & 0.91 (0.92)\\ 
			\bottomrule 
		\end{tabular} 
		
	\end{center}
	\caption{Mean values of Edit Overhead, Correction Time Score and Relative-Correctness. Values in parentheses refer to using truncated samples during training.}
	\label{table:metrics-overview}
\end{table*}

\clearpage

\begin{table*}[h]
	\small
	\begin{center}
		
		\begin{tabular}{l c c c c c c} 
			\toprule 
			\multirow{2}{*}{\bf Task/Model} & \multicolumn{3}{c}{\bf EO } & \multicolumn{3}{c}{\bf EO with prophecies} \\ 
			\cmidrule{2-4} \cmidrule{5-7}
			& \bf $\Delta0$ & \bf $\Delta1$ & \bf $\Delta2$ & \bf $\Delta0$ & \bf $\Delta1$ & \bf $\Delta2$ \\ 
			\midrule
			\sc Sequence tagging & & & & & & \\ 
			\cmidrule{1-1} 
			\hspace{0.2cm}Chunk & & & & & & \\ 
			\hspace{0.4cm} \sc LSTM & 0.00 (0.00) & 0.00 (0.00) & 0.00 (0.00) & 0.00 (0.00) & 0.00 (0.00) & 0.00 (0.00)\\ 
			\hspace{0.4cm} \sc LSTM+CRF & 0.06 (0.03) & 0.01 (0.00) & 0.00 (0.00) & 0.03 (0.03) & 0.00 (0.00) & 0.00 (0.00)\\ 
			\hspace{0.4cm} \sc BiLSTM & 0.10 (0.08) & 0.06 (0.05) & 0.03 (0.03) & 0.09 (0.11) & 0.05 (0.07) & 0.04 (0.05)\\ 
			\hspace{0.4cm} \sc BiLSTM+CRF & 0.16 (0.07) & 0.06 (0.04) & 0.03 (0.02) & 0.10 (0.09) & 0.06 (0.05) & 0.04 (0.03)\\ 
			\hspace{0.4cm} \sc BERT & 0.17 (0.06) & 0.10 (0.03) & 0.09 (0.02) & 0.10 (0.06) & 0.03 (0.03) & 0.03 (0.02)\\ 
			\hspace{0.2cm}Named Entity Recognition & & & & & & \\ 
			\hspace{0.4cm} \sc LSTM & 0.00 (0.00) & 0.00 (0.00) & 0.00 (0.00) & 0.00 (0.00) & 0.00 (0.00) & 0.00 (0.00)\\ 
			\hspace{0.4cm} \sc LSTM+CRF & 0.05 (0.04) & 0.02 (0.01) & 0.00 (0.00) & 0.04 (0.04) & 0.01 (0.01) & 0.00 (0.00)\\ 
			\hspace{0.4cm} \sc BiLSTM & 0.07 (0.06) & 0.04 (0.03) & 0.02 (0.02) & 0.08 (0.08) & 0.05 (0.05) & 0.04 (0.03)\\ 
			\hspace{0.4cm} \sc BiLSTM+CRF & 0.08 (0.06) & 0.04 (0.03) & 0.02 (0.02) & 0.09 (0.08) & 0.05 (0.04) & 0.04 (0.03)\\ 
			\hspace{0.4cm} \sc BERT & 0.17 (0.06) & 0.14 (0.03) & 0.13 (0.02) & 0.08 (0.06) & 0.04 (0.03) & 0.03 (0.02)\\ 
			\hspace{0.2cm}Part-of-Speech Tagging & & & & & & \\ 
			\hspace{0.4cm} \sc LSTM & 0.00 (0.00) & 0.00 (0.00) & 0.00 (0.00) & 0.00 (0.00) & 0.00 (0.00) & 0.00 (0.00)\\ 
			\hspace{0.4cm} \sc LSTM+CRF & 0.05 (0.03) & 0.00 (0.00) & 0.00 (0.00) & 0.03 (0.03) & 0.00 (0.00) & 0.00 (0.00)\\ 
			\hspace{0.4cm} \sc BiLSTM & 0.08 (0.06) & 0.03 (0.02) & 0.02 (0.01) & 0.07 (0.06) & 0.04 (0.03) & 0.03 (0.02)\\ 
			\hspace{0.4cm} \sc BiLSTM+CRF & 0.09 (0.06) & 0.03 (0.03) & 0.02 (0.02) & 0.07 (0.06) & 0.04 (0.03) & 0.03 (0.02)\\ 
			\hspace{0.4cm} \sc BERT & 0.52 (0.05) & 0.48 (0.01) & 0.46 (0.01) & 0.07 (0.05) & 0.02 (0.02) & 0.02 (0.02)\\ 
			\hspace{0.2cm}Semantic Role Labeling & & & & & & \\ 
			\hspace{0.4cm} \sc LSTM & 0.00 (0.00) & 0.00 (0.00) & 0.00 (0.00) & 0.00 (0.00) & 0.00 (0.00) & 0.00 (0.00)\\ 
			\hspace{0.4cm} \sc LSTM+CRF & 0.21 (0.29) & 0.13 (0.23) & 0.09 (0.18) & 0.32 (0.26) & 0.26 (0.21) & 0.21 (0.17)\\ 
			\hspace{0.4cm} \sc BiLSTM & 0.31 (0.37) & 0.26 (0.33) & 0.23 (0.30) & 0.34 (0.40) & 0.31 (0.37) & 0.28 (0.34)\\ 
			\hspace{0.4cm} \sc BiLSTM+CRF & 0.28 (0.31) & 0.24 (0.27) & 0.21 (0.23) & 0.28 (0.35) & 0.25 (0.32) & 0.22 (0.28)\\ 
			\hspace{0.4cm} \sc BERT & 0.43 (0.33) & 0.40 (0.29) & 0.37 (0.25) & 0.22 (0.26) & 0.19 (0.23) & 0.16 (0.20)\\ 
			\hspace{0.2cm}Slot Filling (ATIS) & & & & & & \\ 
			\hspace{0.4cm} \sc LSTM & 0.00 (0.00) & 0.00 (0.00) & 0.00 (0.00) & 0.00 (0.00) & 0.00 (0.00) & 0.00 (0.00)\\ 
			\hspace{0.4cm} \sc LSTM+CRF & 0.02 (0.01) & 0.00 (0.00) & 0.00 (0.00) & 0.01 (0.03) & 0.00 (0.00) & 0.00 (0.00)\\ 
			\hspace{0.4cm} \sc BiLSTM & 0.02 (0.02) & 0.01 (0.01) & 0.00 (0.00) & 0.02 (0.02) & 0.01 (0.01) & 0.00 (0.00)\\ 
			\hspace{0.4cm} \sc BiLSTM+CRF & 0.03 (0.01) & 0.01 (0.00) & 0.00 (0.00) & 0.04 (0.02) & 0.01 (0.01) & 0.00 (0.00)\\ 
			\hspace{0.4cm} \sc BERT & 0.22 (0.03) & 0.15 (0.01) & 0.13 (0.00) & 0.06 (0.03) & 0.00 (0.01) & 0.00 (0.00)\\ 
			\hspace{0.2cm}Slot Filling (SNIPS) & & & & & & \\ 
			\hspace{0.4cm} \sc LSTM & 0.00 (0.00) & 0.00 (0.00) & 0.00 (0.00) & 0.00 (0.00) & 0.00 (0.00) & 0.00 (0.00)\\ 
			\hspace{0.4cm} \sc LSTM+CRF & 0.06 (0.04) & 0.02 (0.01) & 0.01 (0.00) & 0.10 (0.07) & 0.03 (0.02) & 0.02 (0.00)\\ 
			\hspace{0.4cm} \sc BiLSTM & 0.12 (0.08) & 0.07 (0.05) & 0.05 (0.03) & 0.17 (0.13) & 0.10 (0.07) & 0.07 (0.04)\\ 
			\hspace{0.4cm} \sc BiLSTM+CRF & 0.11 (0.08) & 0.05 (0.05) & 0.03 (0.03) & 0.17 (0.14) & 0.09 (0.07) & 0.06 (0.05)\\ 
			\hspace{0.4cm} \sc BERT & 0.37 (0.08) & 0.29 (0.04) & 0.23 (0.02) & 0.12 (0.09) & 0.05 (0.04) & 0.03 (0.02)\\
			& & & & & & \\ 
			\sc Sequence classification & & & & & & \\ 
			\cmidrule{1-1} 
			\hspace{0.2cm}Intent (ATIS) & & & & & & \\ 
			\hspace{0.4cm} \sc LSTM & 0.48 (0.21) & 0.34 (0.13) & 0.26 (0.09) & 0.77 (0.78) & 0.72 (0.72) & 0.65 (0.64)\\ 
			\hspace{0.4cm} \sc LSTM+CRF & - & - & -& - & - & - \\
			\hspace{0.4cm} \sc BiLSTM & 0.40 (0.24) & 0.32 (0.16) & 0.25 (0.12) & 0.66 (0.77) & 0.51 (0.70) & 0.36 (0.63)\\ 
			\hspace{0.4cm} \sc BiLSTM+CRF & - & - & -& - & - & - \\
			\hspace{0.4cm} \sc BERT & 0.49 (0.20) & 0.16 (0.12) & 0.11 (0.09) & 0.38 (0.29) & 0.24 (0.18) & 0.15 (0.13)\\ 
			\hspace{0.2cm}Intent (SNIPS) & & & & & & \\ 
			\hspace{0.4cm} \sc LSTM & 0.31 (0.24) & 0.23 (0.16) & 0.16 (0.11) & 0.57 (0.49) & 0.48 (0.40) & 0.40 (0.33)\\ 
			\hspace{0.4cm} \sc LSTM+CRF & - & - & -& - & - & - \\
			\hspace{0.4cm} \sc BiLSTM & 0.26 (0.23) & 0.22 (0.16) & 0.16 (0.10) & 0.55 (0.49) & 0.48 (0.42) & 0.41 (0.34)\\ 
			\hspace{0.4cm} \sc BiLSTM+CRF & - & - & -& - & - & - \\
			\hspace{0.4cm} \sc BERT & 0.30 (0.22) & 0.25 (0.14) & 0.18 (0.08) & 0.40 (0.35) & 0.31 (0.26) & 0.23 (0.16)\\ 
			\hspace{0.2cm}Positive/Negative & & & & & & \\ 
			\hspace{0.4cm} \sc LSTM & 0.38 (0.40) & 0.31 (0.31) & 0.24 (0.25) & 0.65 (0.68) & 0.59 (0.62) & 0.52 (0.56)\\ 
			\hspace{0.4cm} \sc LSTM+CRF & - & - & -& - & - & - \\
			\hspace{0.4cm} \sc BiLSTM & 0.37 (0.39) & 0.31 (0.31) & 0.27 (0.27) & 0.63 (0.58) & 0.56 (0.52) & 0.48 (0.46)\\ 
			\hspace{0.4cm} \sc BiLSTM+CRF & - & - & -& - & - & - \\
			\hspace{0.4cm} \sc BERT & 0.58 (0.45) & 0.49 (0.35) & 0.41 (0.27) & 0.56 (0.55) & 0.48 (0.48) & 0.41 (0.41)\\ 
			\hspace{0.2cm}Pros/Cons & & & & & & \\ 
			\hspace{0.4cm} \sc LSTM & 0.15 (0.13) & 0.10 (0.08) & 0.07 (0.06) & 0.32 (0.27) & 0.25 (0.20) & 0.19 (0.15)\\ 
			\hspace{0.4cm} \sc LSTM+CRF & - & - & -& - & - & - \\
			\hspace{0.4cm} \sc BiLSTM & 0.14 (0.14) & 0.08 (0.09) & 0.06 (0.06) & 0.32 (0.30) & 0.26 (0.23) & 0.21 (0.17)\\ 
			\hspace{0.4cm} \sc BiLSTM+CRF & - & - & -& - & - & - \\
			\hspace{0.4cm} \sc BERT & 0.20 (0.14) & 0.11 (0.08) & 0.08 (0.06) & 0.24 (0.22) & 0.16 (0.14) & 0.11 (0.10)\\
			\bottomrule 
		\end{tabular}

	\end{center}
	\caption{Mean Edit Overhead and Delay of one or two time steps. Values in parentheses refer to using truncated samples during training.}
	\label{table:eo-delay}
\end{table*}

\clearpage

\begin{table*}[h]
	\small
	\begin{center}
		
		\begin{tabular}{l c c c c c c} 
			\toprule 
			\multirow{2}{*}{\bf Task/Model} & \multicolumn{3}{c}{\bf RC } & \multicolumn{3}{c}{\bf RC with prophecies} \\ 
			\cmidrule{2-4} \cmidrule{5-7}
			& \bf $\Delta0$ & \bf $\Delta1$ & \bf $\Delta2$ & \bf $\Delta0$ & \bf $\Delta1$ & \bf $\Delta2$ \\ 
			\midrule
			\sc Sequence tagging & & & & & & \\ 
			\cmidrule{1-1} 
			\hspace{0.2cm}Chunk & & & & & & \\ 
			\hspace{0.4cm} \sc LSTM& 1.00 (1.00) & 1.00 (1.00) & 1.00 (1.00)& 1.00 (1.00) & 1.00 (1.00) & 1.00 (1.00)\\ 
			\hspace{0.4cm} \sc LSTM+CRF& 0.94 (0.97) & 0.99 (1.00) & 1.00 (1.00)& 0.97 (0.97) & 1.00 (1.00) & 1.00 (1.00)\\ 
			\hspace{0.4cm} \sc BiLSTM& 0.84 (0.89) & 0.91 (0.93) & 0.94 (0.95)& 0.90 (0.87) & 0.94 (0.92) & 0.96 (0.94)\\ 
			\hspace{0.4cm} \sc BiLSTM+CRF& 0.79 (0.92) & 0.93 (0.96) & 0.95 (0.97)& 0.91 (0.91) & 0.94 (0.95) & 0.96 (0.97)\\ 
			\hspace{0.4cm} \sc BERT& 0.67 (0.91) & 0.74 (0.94) & 0.75 (0.95)& 0.87 (0.92) & 0.94 (0.95) & 0.95 (0.96)\\ 
			\hspace{0.2cm}Named Entity Recognition & & & & & & \\ 
			\hspace{0.4cm} \sc LSTM& 1.00 (1.00) & 1.00 (1.00) & 1.00 (1.00)& 1.00 (1.00) & 1.00 (1.00) & 1.00 (1.00)\\ 
			\hspace{0.4cm} \sc LSTM+CRF& 0.95 (0.96) & 0.98 (0.99) & 1.00 (1.00)& 0.96 (0.96) & 0.99 (0.99) & 1.00 (1.00)\\ 
			\hspace{0.4cm} \sc BiLSTM& 0.91 (0.93) & 0.95 (0.95) & 0.96 (0.97)& 0.91 (0.92) & 0.95 (0.95) & 0.96 (0.96)\\ 
			\hspace{0.4cm} \sc BiLSTM+CRF& 0.92 (0.94) & 0.95 (0.97) & 0.97 (0.98)& 0.92 (0.93) & 0.95 (0.96) & 0.97 (0.97)\\ 
			\hspace{0.4cm} \sc BERT& 0.49 (0.93) & 0.51 (0.96) & 0.53 (0.97)& 0.90 (0.93) & 0.94 (0.96) & 0.96 (0.97)\\ 
			\hspace{0.2cm}Part-of-Speech Tagging & & & & & & \\ 
			\hspace{0.4cm} \sc LSTM& 1.00 (1.00) & 1.00 (1.00) & 1.00 (1.00)& 1.00 (1.00) & 1.00 (1.00) & 1.00 (1.00)\\ 
			\hspace{0.4cm} \sc LSTM+CRF& 0.95 (0.96) & 1.00 (1.00) & 1.00 (1.00)& 0.97 (0.97) & 1.00 (1.00) & 1.00 (1.00)\\ 
			\hspace{0.4cm} \sc BiLSTM& 0.89 (0.93) & 0.95 (0.96) & 0.96 (0.98)& 0.92 (0.93) & 0.95 (0.96) & 0.97 (0.98)\\ 
			\hspace{0.4cm} \sc BiLSTM+CRF& 0.89 (0.92) & 0.95 (0.96) & 0.97 (0.97)& 0.92 (0.93) & 0.96 (0.96) & 0.97 (0.97)\\ 
			\hspace{0.4cm} \sc BERT& 0.30 (0.93) & 0.30 (0.97) & 0.29 (0.97)& 0.90 (0.93) & 0.96 (0.96) & 0.97 (0.97)\\ 
			\hspace{0.2cm}Semantic Role Labeling & & & & & & \\ 
			\hspace{0.4cm} \sc LSTM& 1.00 (1.00) & 1.00 (1.00) & 1.00 (1.00)& 1.00 (1.00) & 1.00 (1.00) & 1.00 (1.00)\\ 
			\hspace{0.4cm} \sc LSTM+CRF& 0.71 (0.83) & 0.83 (0.87) & 0.89 (0.91)& 0.82 (0.85) & 0.87 (0.89) & 0.90 (0.92)\\ 
			\hspace{0.4cm} \sc BiLSTM& 0.59 (0.60) & 0.66 (0.65) & 0.70 (0.69)& 0.62 (0.62) & 0.66 (0.66) & 0.70 (0.70)\\ 
			\hspace{0.4cm} \sc BiLSTM+CRF& 0.65 (0.72) & 0.72 (0.76) & 0.76 (0.80)& 0.71 (0.72) & 0.75 (0.76) & 0.79 (0.79)\\ 
			\hspace{0.4cm} \sc BERT& 0.25 (0.70) & 0.27 (0.74) & 0.28 (0.77)& 0.69 (0.67) & 0.74 (0.72) & 0.77 (0.75)\\ 
			\hspace{0.2cm}Slot Filling (ATIS) & & & & & & \\ 
			\hspace{0.4cm} \sc LSTM& 1.00 (1.00) & 1.00 (1.00) & 1.00 (1.00)& 1.00 (1.00) & 1.00 (1.00) & 1.00 (1.00)\\ 
			\hspace{0.4cm} \sc LSTM+CRF& 0.98 (0.98) & 1.00 (1.00) & 1.00 (1.00)& 0.99 (0.97) & 1.00 (1.00) & 1.00 (1.00)\\ 
			\hspace{0.4cm} \sc BiLSTM& 0.97 (0.98) & 0.99 (0.99) & 1.00 (1.00)& 0.97 (0.98) & 0.99 (0.99) & 1.00 (1.00)\\ 
			\hspace{0.4cm} \sc BiLSTM+CRF& 0.97 (0.98) & 0.99 (0.99) & 1.00 (1.00)& 0.95 (0.98) & 0.99 (0.99) & 1.00 (1.00)\\ 
			\hspace{0.4cm} \sc BERT& 0.56 (0.97) & 0.65 (0.99) & 0.71 (0.99)& 0.93 (0.97) & 0.99 (0.99) & 1.00 (0.99)\\ 
			\hspace{0.2cm}Slot Filling (SNIPS) & & & & & & \\ 
			\hspace{0.4cm} \sc LSTM& 1.00 (1.00) & 1.00 (1.00) & 1.00 (1.00)& 1.00 (1.00) & 1.00 (1.00) & 1.00 (1.00)\\ 
			\hspace{0.4cm} \sc LSTM+CRF& 0.93 (0.96) & 0.98 (0.99) & 0.99 (1.00)& 0.90 (0.93) & 0.97 (0.99) & 0.99 (1.00)\\ 
			\hspace{0.4cm} \sc BiLSTM& 0.84 (0.90) & 0.91 (0.93) & 0.94 (0.96)& 0.81 (0.85) & 0.88 (0.91) & 0.92 (0.94)\\ 
			\hspace{0.4cm} \sc BiLSTM+CRF& 0.88 (0.91) & 0.94 (0.95) & 0.96 (0.96)& 0.82 (0.86) & 0.90 (0.93) & 0.93 (0.95)\\ 
			\hspace{0.4cm} \sc BERT& 0.41 (0.91) & 0.50 (0.95) & 0.60 (0.97)& 0.86 (0.90) & 0.94 (0.95) & 0.97 (0.97)\\ 
			& & & & & & \\ 
			\sc Sequence classification & & & & & & \\ 
			\cmidrule{1-1} 
			\hspace{0.2cm}Intent (ATIS) & & & & & & \\ 
			\hspace{0.4cm} \sc LSTM& 0.77 (0.91) & 0.84 (0.95) & 0.89 (0.96)& 0.52 (0.54) & 0.58 (0.59) & 0.64 (0.66)\\ 
			\hspace{0.4cm} \sc LSTM+CRF& - & - & - & - & - & - \\ 
			\hspace{0.4cm} \sc BiLSTM& 0.85 (0.90) & 0.88 (0.94) & 0.91 (0.95)& 0.70 (0.50) & 0.77 (0.55) & 0.85 (0.61)\\ 
			\hspace{0.4cm} \sc BiLSTM+CRF& - & - & - & - & - & - \\ 
			\hspace{0.4cm} \sc BERT& 0.84 (0.91) & 0.93 (0.94) & 0.95 (0.96)& 0.88 (0.90) & 0.92 (0.93) & 0.95 (0.96)\\ 
			\hspace{0.2cm}Intent (SNIPS) & & & & & & \\ 
			\hspace{0.4cm} \sc LSTM& 0.85 (0.90) & 0.89 (0.93) & 0.92 (0.95)& 0.71 (0.77) & 0.75 (0.80) & 0.80 (0.84)\\ 
			\hspace{0.4cm} \sc LSTM+CRF& - & - & - & - & - & - \\ 
			\hspace{0.4cm} \sc BiLSTM& 0.86 (0.91) & 0.89 (0.93) & 0.91 (0.95)& 0.71 (0.76) & 0.75 (0.80) & 0.80 (0.84)\\ 
			\hspace{0.4cm} \sc BiLSTM+CRF& - & - & - & - & - & - \\ 
			\hspace{0.4cm} \sc BERT& 0.83 (0.91) & 0.86 (0.94) & 0.91 (0.96)& 0.83 (0.86) & 0.87 (0.90) & 0.90 (0.94)\\ 
			\hspace{0.2cm}Positive/Negative & & & & & & \\ 
			\hspace{0.4cm} \sc LSTM& 0.78 (0.79) & 0.81 (0.82) & 0.84 (0.85)& 0.74 (0.72) & 0.76 (0.75) & 0.79 (0.77)\\ 
			\hspace{0.4cm} \sc LSTM+CRF& - & - & - & - & - & - \\ 
			\hspace{0.4cm} \sc BiLSTM& 0.82 (0.83) & 0.84 (0.85) & 0.86 (0.87)& 0.76 (0.77) & 0.79 (0.79) & 0.82 (0.81)\\ 
			\hspace{0.4cm} \sc BiLSTM+CRF& - & - & - & - & - & - \\ 
			\hspace{0.4cm} \sc BERT& 0.64 (0.80) & 0.66 (0.84) & 0.68 (0.86)& 0.79 (0.79) & 0.82 (0.82) & 0.84 (0.84)\\ 
			\hspace{0.2cm}Pros/Cons & & & & & & \\ 
			\hspace{0.4cm} \sc LSTM& 0.93 (0.94) & 0.95 (0.96) & 0.96 (0.97)& 0.88 (0.90) & 0.90 (0.92) & 0.93 (0.94)\\ 
			\hspace{0.4cm} \sc LSTM+CRF& - & - & - & - & - & - \\ 
			\hspace{0.4cm} \sc BiLSTM& 0.94 (0.94) & 0.96 (0.96) & 0.97 (0.97)& 0.85 (0.88) & 0.87 (0.91) & 0.90 (0.93)\\ 
			\hspace{0.4cm} \sc BiLSTM+CRF& - & - & - & - & - & - \\ 
			\hspace{0.4cm} \sc BERT& 0.91 (0.94) & 0.94 (0.96) & 0.95 (0.97)& 0.91 (0.92) & 0.94 (0.94) & 0.95 (0.96)\\ 
			\bottomrule 
		\end{tabular} 
		
	\end{center}
	\caption{Mean Relative Correctness and Delay of one or two time steps. Values in parentheses refer to using truncated samples during training.}
	\label{table:rc-delay}
\end{table*}

\end{document}